\documentclass[10pt,oneside,article]{memoir} 

\usepackage{ragged2e} 
\usepackage{textcase}

\usepackage{float}

\setstocksize{297mm}{210mm}
\settrimmedsize{\stockheight}{195mm}{*}
\settypeblocksize{671.6pt}{335.8pt}{*}

\setlrmarginsandblock{2cm}{*}{1}
\setulmarginsandblock{2cm}{*}{1}
\setheadfoot{1cm}{\footskip}  

\checkandfixthelayout[nearest] 

\usepackage{varwidth}

\usepackage{microtype}

\usepackage{tipa}

\usepackage{moreverb,url}

\usepackage[colorlinks,bookmarksopen,bookmarksnumbered,citecolor=red,urlcolor=red,breaklinks=true]{hyperref}

\usepackage{crimson}
\usepackage{carlito}

\usepackage{tipx}
\usepackage{pifont}

\usepackage{array}
\usepackage{supertabular}
\usepackage{tabularray}
\usepackage{hhline}
\usepackage{multirow}
\usepackage{graphicx}
\usepackage{vcell}
\usepackage{eurosym}

\usepackage{iftex,amsmath,amsthm}
\usepackage{amssymb,amsfonts,textcomp}

\usepackage{enumitem}

\usepackage{hepnames}

\usepackage[T1]{fontenc} 
\usepackage{junicode,newtxtext,newtxmath} 

\usepackage{xpatch}
\makeatletter
\xpatchcmd\@HepGenStyle{\mathit}{\symit}{}{\show\fail}
\makeatother

\usepackage[english,french]{babel}

\usepackage[compact]{titlesec}

\usepackage[normalem]{ulem}

\usepackage{graphicx, caption, subfig, xcolor}
\definecolor{darkblue}{rgb}{0, 0, 0.5}
\hypersetup{colorlinks=true,citecolor=darkblue, linkcolor=darkblue, urlcolor=darkblue}
\usepackage{colortbl}

\usepackage{tikz}
\usetikzlibrary{fit}

\usepackage{adjustbox}

\makepagestyle{mypagestyle} 
\copypagestyle{mypagestyle}{empty}
\makeheadrule{mypagestyle}{\textwidth}{\normalrulethickness}
\makeoddfoot{mypagestyle}{}{\thepage~of~\thelastpage}{} 

\usepackage{enumerate}

\usepackage[output-decimal-marker={,}]{siunitx}

\DeclareMathOperator{\op}{\mathit{op}}

\newtheoremstyle{property}
{3pt}
{3pt}
{}
{1cm}
{}%
{:}%
{.5em}%
{}%

\theoremstyle{plain}

\newtheoremstyle{note}
{1em}
{1em}
{\small\rmfamily}
{}
{\color{MyDarkRed}\small\bfseries\rmfamily}
{}
{.5em}
{}

\newtheoremstyle{example}
{\parskip}
{\parskip}
{}
{}
{\bfseries}
{}
{\parskip}
{}

\newtheoremstyle{theoremdd}
{\topsep}
{\topsep}
{\itshape}
{0pt}
{\bfseries}
{. ---}
{ }
{\thmname{#1}\thmnumber{ #2}\textnormal{\thmnote{ (#3)}}}

\theoremstyle{theoremdd}
\newtheorem{thmd}{Theorem}[section]

\newtheorem{lemma}{Lemma}[section]

\theoremstyle{note}

\theoremstyle{example}

\counterwithout{section}{section}

\newenvironment{titlepage}{\hfill\par\noindent\trivlist\item\ignorespaces\Large}%
{\endtrivlist}

\newenvironment{affiliation}{\noindent\trivlist\item\ignorespaces}%
{\par\noindent\hfill\endtrivlist}

\newcommand{\keywords}[1]{{\fontsize{9}{10}\selectfont\textbf{\textit{Keywords---}} \addfontfeature{LetterSpace=1.1}\hspace{1mm}{#1}\allowbreak\par\hfill}}

\addto\captionsenglish{}

\usepackage[round,authoryear,sort&compress]{natbib}

\makeatletter

\newif\ifbrief
\newif\ifpubrec

\newskip\listtopsepi
\newskip\listtopsepii
\newskip\listtopsepiii
\newskip\listitemsepi
\newskip\listitemsepii
\newskip\listitemsepiii
\newlength\listleftmargini
\newlength\listleftmarginii
\newlength\listleftmarginiii
\newlength\listlabelwidthi
\newlength\listlabelwidthii
\newlength\listlabelwidthiii
\newlength\listlabelsepi
\newlength\listlabelsepii
\newlength\listlabelsepiii
\newlength\listrightmargin
\newcount\listdepth
\def\nomainitem{\global\@nomainitemtrue}
\newlength\lpalignmaxwd
\def\lpalign#1{\global\@lpaligntrue\global\settowidth\lpalignmaxwd{#1}}
\ifbrief
\else
\fi
\newskip\topsepii \topsepii2pt
\newskip\topsepiii \topsepiii2pt
\newskip\itemsepi \itemsepi0pt
\newskip\itemsepii \itemsepii0pt
\newskip\itemsepiii \itemsepiii0pt
\newdimen\LabelSep \LabelSep4.7pt
\def\@listI{\leftmargin\leftmargini
\labelwidth\leftmargini
\advance\labelwidth-\labelsep
\parsep 0\p@%
\topsep \topsepi
\itemsep\itemsepi}%
\def\@listii {\leftmargin\leftmarginii
\labelwidth\leftmarginii
\advance\labelwidth-\labelsep
\topsep\topsepii
\parsep 0pt
\itemsep\itemsepii}
\def\@listiii {\leftmargin\leftmarginiii
\labelwidth\leftmarginiii
\advance\labelwidth-\labelsep
\topsep\topsepiii
\parsep 0pt
\itemsep\itemsepiii}
\def\@listiv {\leftmargin\leftmarginiv
\labelwidth\leftmarginiv
\advance\labelwidth-\labelsep}
\def\@listv  {\leftmargin\leftmarginv
\labelwidth\leftmarginv
\advance\labelwidth-\labelsep}
\def\@listvi {\leftmargin\leftmarginvi
\labelwidth\leftmarginvi
\advance\labelwidth-\labelsep}
\font\lcir = lcircle10 at 12pt
\newcommand\bulls{\hbox{\lcir\char'162}}
\def\textbullet{\leavevmode\raise3.5pt\bulls\hskip-2pt}
\def\textasteriskcentered{\leavevmode\raise-1.5pt\hbox{*}}
\def\textperiodcentered{\leavevmode\raise1.5pt\hbox{\bulls}}
\newenvironment{alphalist}{%
\begin{enumerate}%
}{%
\end{enumerate}}

\newcounter{eqnnosave}          
\newenvironment{exlist}{
\listleftmargini 36pt
\listlabelwidthi 30pt
\listlabelsepi 6pt
\listitemsepi12pt

\begin{enumerate}%
\setcounter{enumi}{\arabic{eqnnosave}}
}%
{\end{enumerate}%
\setcounter{eqnnosave}{\arabic{enumi}}%
}

\newcounter{deflist}[newdeflist]
\newcounter{subdeflist}[deflist]

\@addtoreset{deflist}{newdeflist}

\newenvironment{deflist}{\protected@edef\@currentlabelname{}\par\list{}{\leftmargin0pt\rightmargin18pt\itemindent-\leftmargin\labelwidth\z@\labelsep\z@\itemsep6pt\topsep12pt plus2pt
}\raggedright}{\endlist}

\renewcommand\paragraph{\@startsection{paragraph}{4}{\z@}%
{-2.5ex\@plus -1ex \@minus -.25ex}%
{1.25ex \@plus .25ex}%
{\normalfont\normalsize\bfseries}}

\renewcommand\subparagraph{\@startsection{subparagraph}{5}{\z@}%
{-2.5ex\@plus -1ex \@minus -.25ex}%
{1.25ex \@plus .25ex}%
{\normalfont\normalsize\bfseries}}

\renewcommand{\thefigure}{\arabic{chapter}.\alph{figure}}

\newcommand{\leqnomode}{\tagsleft@true\let\veqno\@@leqno}
\newcommand{\reqnomode}{\tagsleft@false\let\veqno\@@eqno}

\makeatother

\begin{document}

\begin{otherlanguage}{english}
\begin{Spacing}{1.1}

\begin{titlepage}\thispagestyle{empty}
A theoretical model of dynamical grammatical gender shifting based on set-valued set function\end{titlepage}



\begin{affiliation}Mohamed El Idrissi, ENS Paris-Saclay, mohamed.el\_idrissi@ens-paris-saclay.fr
\end{affiliation}

\begin{abstract}\thispagestyle{empty}

\noindent
This study investigates the diverse characteristics of nouns, focusing on both semantic (e.g., countable/uncountable) and morphosyntactic (e.g., masculine/feminine) distinctions. We explore inter-word variations for gender markers in noun morphology. Grammatical gender shift is a widespread phenomenon in languages around the world. The aim is to uncover, through a data-driven formal model, the underlying patterns governing the variation of lexemes. To this end, we propose a new computational component dedicated to pairing items with morphological templates (e.g., the result of a generated item-template pair: $(\textit{funas}, \{\mathit{N}, +\mathit{SG}, -\mathit{PL}, -\mathit{M}, +\mathit{F}, -\mathit{COL}, +\mathit{SING}\})$, with its spell-out form: \textit{ða-funast} ‘cow’). This process is formally represented by the Template-Based and Modular Cognitive model. This proposed model, defined by a set-valued set function $h:$ $\mathcal{P}(M)$ $\rightarrow$ $\mathcal{P}(M)$, predicts the non-linear dynamical mapping of lexical items onto morphological templates. By applying this formalism, we present a unified framework for understanding the complexities of morphological markings across languages. Through empirical observations, we demonstrate how these shifts, as well as non-gender shifts, arise during lexical changes, especially in Riffian. Our model posits that these variant markings emerge due to template shifts occurring during word and meaning formation. This study achieves two primary objectives. First, on the formal side, we prove the model's representational completeness in learning and prediction. Second, on the linguistic side, we challenge and broaden the conventional view of word formation by formally demonstrating that conversion is applicable to noun-to-noun derivation. This mathematical model not only contributes to a deeper understanding of morphosyntactic variation but also offers potential applications in other fields requiring precise modelling of linguistic patterns.
\end{abstract}

\keywords{morphosyntactic theory, semantic theory, mathematical formalism, cognitive linguistics, artificial intelligence, computational linguistics, corpus linguistics}

\hfill

\section{Introduction}\label{intro}

Synthetic languages assign inflectional markers to lexical items, but these markings vary across words. Depending on semantic properties, the radical\footnote{A radical, also named base by some researchers \citep[p.311]{Chelliah2010Handbook}, is a distinct linguistic unit that differs from a root. This should be viewed in contrast to inflectional morphemes. A root or stem morpheme, along with thematic or scheme, and derivational morphemes constitute the radical.} of nouns can be paired either with all the elements of grammatical paradigms (e.g., in French: \textit{le lion} ‘the lion’, \textit{la lionne} ‘the lioness’, \textit{les lions} ‘the lions’,  \textit{les lionnes} ‘the lionesses’)\footnote{The following abbreviations are employed throughout this paper: PL (Plural), SG (Singular), M (Masculine), F (Feminine), V (Verb), N (Noun), NP (Noun Phrase), SING (Singulative), COL (Collective), DEF (Definite), and CONV (Conversion).}  or only with some of them. Thus, non-animate countable nouns do not inflect in grammatical gender (e.g., in French: \textit{la valise} ‘the suitcase.F’,  \textit{les valises} ‘the suitcases.F’, but not \textit{*le valise}). Likewise, uncountable nouns, which also bear the non-animate meaning, exhibit no gender or number-based inflectional variation (e.g., in French: \textit{le beurre} ‘butter.M.SG’, but not \textit{*la beurre}, \textit{*les beurres}). In addition, a particular marker of a grammatical paradigm, especially gender, is not univocally tied to a single semantic class. Hence, despite lacking inflection, those nouns may be marked as masculine or feminine, as in Riffian\footnote{Riffian is a Berber language spoken in the Rif mountains, located north of Morocco. Approximately 3 to 4 million people speak Riffian. For those seeking an in-depth exploration of the Riffian language, the following publications offer valuable insights and comprehensive coverage \citep{lafkioui_atlas_2007,cadi_systeme_1987}.} or French (see Table \ref{tab1}).

\begin{table}[h!]
  \centering
  \caption{Morphosyntactic gender contrasts in non-animate versus uncountable nouns}
  \resizebox{15cm}{!}{%
    \begin{tblr}{
      width = \linewidth,
      colspec = {Q[200]Q[119]Q[173]Q[188]Q[33]Q[184]Q[230]},
      cell{1}{2} = {c=3}{0.48\linewidth,c},
      cell{1}{5} = {c=3}{0.325\linewidth,c},
      cell{3}{1} = {r=2}{t},
      cell{6}{1} = {r=2}{t},
      hline{1} = {1}{black},
      hline{1} = {2}{1}{black},
      hline{1} = {2-8}{},
      hline{1} = {2}{2-8}{},
      hline{2-3} = {2-8}{black},
      hline{2-3} = {2}{2-8}{black},
      hline{8} = {-}{black},
      hline{8} = {2}{-}{black},
      }
	~   & French~    &       ~            &     ~       &  Riffian~ &                &  ~   \\
      ~   & Gender~      &   ~      & ~         & ~  &     ~   & ~ \\
      Non-animate ~ & M & \textit{le port}   & port~      & ~    & \textipa{\textit{a-X:am}}~           & chamber~ \\
        ~    & F     & \textit{la valise}  & suitcase~ & ~     & \textipa{\textit{Ta-zeQbutS}} & crossbody bag~ \\
      ~    & ~           &  ~        & ~         & ~ & ~            & ~ \\
      Uncountable ~ & M       & \textit{le beurre} & butter~   & ~      & \textipa{\textit{a-nzaR}}       & rain~ \\
       ~    & F & \textit{la farine}   & flour~    & ~   & \textipa{\textit{Ta-Rgazt}}        & courage~
    \end{tblr}
  } \label{tab1}
\end{table}

As one can see, the same gender marker can apply to multiple noun categories. We designate such a phenomenon \textsc{inter-word} variations. Such a gender variation can be predicted for some types of nouns. Those groups are characterised by the word formations they undergo and their semantic features. A clear example comes from French, where uncountable nouns formed from adjectives with the suffix \textit{-té}, such as \textit{beau} ‘beautiful’, are always assigned feminine gender, as in \textit{la beauté} ‘beauty.F’ \citep{Gruaz1988}.

When such a process is triggered, the derived lexeme may obtain different grammatical markers from the base lexeme. It is generally accepted that this phenomenon can occur during \textsc{inter-lexical derivations}\footnote{Where necessary for scientific clarity, new conceptual terms, marked in small capitals for emphasis, have been introduced and we have provided their definitions.}\footnote{Intra-lexical derivation is a word formation process that maintains the syntactic category of the base word (e.g., noun to noun), whereas inter-lexical derivation changes the syntactic category (e.g., verb to noun).}, but there is less awareness of this situation with \textsc{intra-lexical derivations} \citep{bauer2005conversion}, where the same phenomenon is observed. Thereby, when a noun derives from another one, a \textsc{morphosyntactic shift}, as we may name such a particularity, can result (see (\ref{problem})).


\begin{exlist}
  {\small
    \item 
    \begin{alphalist}[parsep=0.3mm,itemsep=2mm]

      \begin{otherlanguage}{english}

        \item Countable noun from Countable noun

        In Riffian: \textit{t-fust} ‘bouquet.F’, from \textit{fus} ‘hand.M’

        \item Countable noun from Countable noun

        In French: \textit{le gland} ‘acorn.M’, from \textit{la glande} ‘gland.F’

      \end{otherlanguage}
    \label{problem}
    \end{alphalist}
  }
  
\end{exlist}


In Riffian as well as in French, conversion is employed in (\ref{problem}) to form the derived nouns in question. As a result, the gender of the resultant lexemes is contrary to that of its base lexeme. The use of this word formation provides good evidence that the morphosyntactic shift does not depend on another element (e.g., a derivational marker). Thus, the output is solely determined by inherent properties included in the lexical and grammatical morphemes.

In relation to the previous observation, it is also important to note another major linguistic fact regarding morphosyntactic shift. In (\ref{problem}), the only grammatical marker shifting is gender; however, under these circumstances more than one morpheme can also be subject to change. While such grouped modifications are commonly attested and acknowledged in the case of inter-lexical derivations, we must recognise that intra-lexical derivations similarly show the same specificities (see (\ref{problem2})).


\begin{exlist}
  {\small
    \item
    \begin{alphalist}[parsep=0.3mm,itemsep=2mm]

      \begin{otherlanguage}{english}
      
        \item Countable noun from Uncountable noun

        In Riffian: $\textipa{\textit{Ta-ReS:int}}$ ‘an orange’, from $\textipa{\textit{ReS:in}}$ ‘oranges’

         \item Proper noun from Proper noun

        In French:  \textit{la chiraquie} ‘circle of people associated with Jacques Chirac’, from \textit{Chirac} ‘Chirac (the former French president Jacques Chirac)’

      \end{otherlanguage}
    \label{problem2}
    \end{alphalist}
  }
  
\end{exlist}

In both examples in (\ref{problem2}), the derived nouns acquire two new meaning features that were not present in the base nouns. The base nouns convey collective (see \cite{ElIdrissi2024} for parallels between proper and uncountable nouns) and masculine meanings. In Riffian, the derived noun expresses singulative and feminine meanings, while in French, it is marked by the definite article and feminine gender. As demonstrated through these instances, the scope of shifts is not necessarily a single grammatical unit, but can be a set of semantic elements. Among these elements, gender is often a common target.

The aforementioned scientific conundrum emphasises the intricate theoretical questions surrounding the study of morphological phenomena. Our objective is to propose a model that reflects the chaotic and non-linear\footnote{The concepts of chaotic and non-linear behaviour originate in dynamical systems theory. We argue that analogous patterns in linguistics may be compared to phenomena in the physical or behavioural sciences.} behaviour of grammatical marking. Thus, this study seeks to advance knowledge about the nature of lexemes, particularly their morphological and semantic properties. We will restrict ourselves to the study of grammatical gender shift and item-template pairing. Therefore, we will not address morpheme pairing (or morphosyntactic semiotic encoding), which we consider a separate component of the linguistic system, distinct from item-template pairing. In addition, other grammatical features will not be analysed for the sake of clarity and space.

A theoretical framework, grounded in computational mathematics, is developed to account for these phenomena attested cross-linguistically. The theoretical contribution concerns the formalisation of the learning architecture and its mathematical properties, whereas the model is derived from and evaluated against empirical, data-driven evidence. This study posits that synthetic languages exhibit a shared structural pattern, with Riffian serving as the principal language of analysis. To understand the underlying patterns of phenomena, data with significant variation is preferable. Riffian, with its rich linguistic variation, provides a valuable context for studying how universal principles operate across different languages.

To capture these linguistic facts outlined earlier, occurring whether at the \textsc{inter-word} level (see Table \ref{tab1}) or involving \textsc{morphosyntactic shifts} (see examples (\ref{problem}) and (\ref{problem2})), the Template-Based and Modular Cognitive model (hereinafter TBMC model) is used to this end. This model is a formal mathematical framework based on set-valued set functions. Through the TBMC model, it becomes possible to operate in single-valued fashion and map lexical items to their corresponding morphological templates. The phenomena in question are widespread across diverse language families, to name a few, such as Indo-European, Afro-Asiatic, and Native American and have been approached from multiple theoretical perspectives. Thus, semantics-driven theories have been proposed to account for those morphological variations, but their explanatory power is constrained by language-specific and category-specific (evaluative, uncountable, etc.) limitations. Their focus is on providing semantic justifications for existing morphological forms rather than reconciling them under a unified framework.

While different, these approaches converge on a shared conceptual foundation, which we refer to as the \textsc{faithfulness presumption}. We describe theories as semantics-driven when they prioritise a faithful correspondence between the input (semantics) and the output (morphemes). With this global perspective in mind, two approaches can be found in the literature regarding grammatical gender shift. The first treats any instance of this phenomenon as the result of morphological derivation, even when no obvious markers can be identified (e.g., from F to M, where M is represented by a null marker). Under this view, \textit{le gland}/\textit{la glande} and \textit{la chiraquie}/\textit{Chirac} are analysed as composed of derivational markers. The second, less popular approach assumes that conversion is involved when no derivational marker is bound to the radical (as in \textit{le gland}/\textit{la glande}) ---without, however, explaining the mechanisms underlying this process; while allowing that other cases of gender shift may result from derivational markers (as in \textit{la chiraquie}/\textit{Chirac} where \textit{-i} is a feminine derivational suffix). In this study, by contrast, we propose a third approach. We break down the problem into a two-step dynamic model. We argue that, during any word and meaning formation, the latter solely manages the creation of items and does not handle grammatical meanings. Grammatical meanings are added later by a separate computational component, namely the item-template pairing.

Thus, to model these phenomena, an alternative proposal is considered. A core tenet of our theory is that word forms are organised according to pre-syntactic templates encoded in the lexicon. This model posits that contrastive markings are products of template shifts that occur during word and meaning formation. To gain a comprehensive understanding of grammatical gender shifts, we will systematically examine their characteristics and formal representations. In addition, for the sake of demonstration, it is necessary to focus our analysis on specific types of word and meaning formation, namely conversion and semantic widening, before synthesising our findings.

The remainder of this paper is structured around dual research objectives: advancing our cognitive linguistic understanding of grammatical shifts through the lens of gender, and strengthening the mathematical foundation of the TBMC model via formal demonstration and empirical statistical analysis. From a cognitive standpoint, a major result of this study is the formal demonstration that certain intra-lexical changes can be accounted for as a conversion process within our framework. On the mathematical side, our most significant contribution is establishing the model's representational completeness in both learning and prediction.

The paper is structured as follows: Section 2 provides an overview of related work in linguistics and mathematics. Section 3 presents the theoretical foundation of our study and applies it to Riffian data, focusing on morphosyntactic gender patterns. This section also discusses key properties of our model by determining initial templates through a data-driven method and demonstrating how it learns to predict grammatical templates. It further addresses broader linguistic and cognitive questions. The paper concludes with a summary in Section 4.

\section{Literature review}\label{review}

There is a substantial body of literature addressing grammatical gender shift across various languages. However, these studies primarily offer a static analysis, focusing primarily on the properties that distinguish shifted from unshifted words and that fall outside the scope of the item-template pairing component (see Section \ref{review_cog}) ---such as morphosyntactic structure, semantic features, and so on (see for example \cite{mathieu2018gender,Steriopolo2023,kucerova2022underspecification}). The underlying shifting operation that drives this change is generally overlooked. Investigating this dynamic process constitutes a key novelty of our study. Consequently, this traditional descriptive approach will not be the main focus of our literature review, as it does not address the dynamic and computational aspects of grammatical shifting. Nevertheless, in Section \ref{shift}, we will review studies that treat derivational markers as carrying gender features, given their relevance to our approach. The ensuing sections provide a comprehensive overview of relevant literature, emphasising theoretical morphology and the Riffian nominal system, with a focus on gender.

\subsection{Traditional and novel perspectives on the definite article and Riffian gender}\label{rif_morphology}

As a synthetic and fusional language, Riffian primarily constructs its lexemes through derivation and inflection. As in many synthetic languages, both processes are realised through concatenative (e.g., affixation or null markers) or non-concatenative methods (e.g., gemination). The general structure of a nominal base (nouns and adjectives only), represented by the symbol $⨂$, comprises the following components:

\hfill

{\centering
  \fbox{%
    \parbox{8cm}{%
    \centering
      \vskip 0.2cm
      $\bigotimes$ = countability $+$ radical $+$ gender $+$ number
      \vskip 0.15cm
    }%
  }\par
}

\hfill

This morphological system also presents specific challenges in the areas of morphophonology, morphosyntax and morphosemantics. But, given the limitations of this paper, a thorough exploration of Riffian morphology is beyond the present study's purview. We will only highlight some relevant particularities. Each component constitutes a grammatical paradigm including different meanings in binary oppositions: countability opposes singulative and collective, gender contrasts masculine and feminine, while number differentiates between singular and plural.

However, not all of these markers have spell-out forms. The collective, masculine, and singular are null markers. Hence, if an item gets all these null markers, the radical and the word-form have the same form (e.g., \textipa{\textit{muDRɾus}} ‘slim’ from the item-template pair\footnote{Throughout this article, items are presented with their spell-out form for simplicity; however, they should be viewed as either form-meaning pairs or sets of meanings (see Figure \ref{fig0}). Furthermore, formally, an item refers to an element of the set $W$ (see Section \ref{method}). Although in this study an item denotes a radical morpheme, we prefer not to restrict the definition of item to a specific type of linguistic fact for two reasons. First, a grammatical template can also be paired with other linguistic units, particularly in cases of nominalisation (e.g., in French, \textit{le “en même temps” de Macron} ‘The “at the same time” of Macron’). Second, the TBMC model is not intended to handle only linguistic facts, since the elements of $W$ and  $T$ may refer to any type of entity.}: (\textipa{\textit{muDRus}},  $\{N,+SG,$$-PL,+M,$$-F,+COL,$ $-SING$$\})$). In contrast, the other markers are paired with phonological elements. The singulative exhibits two forms conditioned by number: \textit{-a} in singular and \textit{-i} in plural. As for the plural, the main marker is \textit{-n} for the external plural and there are various forms for the internal plural (see \cite{lafkioui_atlas_2007} for further details).

The grammatical gender is pronounced either as a voiced dental \textit{ð} or voiceless dental \textit{t/θ} depending on phonological environments and dialectal variations. We would like to add some remarks about this morpheme. Notably, the same morpheme may recur multiple times, particularly the singular feminine marker, which often appears at both the initial and final positions of the radical (refer to the nominal inflection in Table \ref{tab2}). It is generally assumed, in Riffian linguistics, that these phonemes form a single morpheme, named a circumfix in linguistics \citep{Tahiri_2022}.

However, our theoretical position in this matter is that they are two identical but separate affixes. This system shares common characteristics with the nominal systems of some Indo-European languages (e.g., in Spanish, the feminine marker \textit{-a} appears twice: \textit{l-a vac-a} ‘cow’). The only difference between these languages and Riffian is that, instead of being affixed to a definite article, one of the feminine markers is prefixed to the countability markers.

\begin{table}[h!]
  \centering
  \caption{Nominal inflection for number and gender}
  \resizebox{13cm}{!}{%
    \begin{tblr}{
      width = 5cm,
      colspec = {Q[7.8cm]Q[2.4cm]Q[2.4cm]},
      hline{1,2,6} = {-}{black},
      hline{1,2,6} = {2}{-}{black},
        }
  Item-template pair   &  Spell-out form      & Gloss~ \\
   (\textit{qzin},  $\{N,+SG,-PL,+M,-F,-COL,+SING\}$) &  \textit{aqzin~}      & dog.SG.M~ \\
      (\textit{qzin},  $\{N,+SG,-PL,-M,+F,-COL,+SING\}$) &  \textit{ð-aqzin-t~}  & dog.SG.F~ \\
     (\textit{qzin},  $\{N,-SG,+PL,+M,-F,-COL,+SING\}$) & \textit{iqzin-en~}   & dog.PL.M~ \\
      (\textit{qzin},  $\{N,-SG,+PL,-M,+F,-COL,+SING\}$) & \textit{ð-iqzin-in~} & dog.PL.F~
    \end{tblr}
  } \label{tab2}
\end{table}

This proposal is based on evidence showing that they are dissociated units. We observed that either could be spelled out separately. Therefore, some nouns can ignore the final feminine marker and use the initial one (e.g., \textipa{\textit{D-aRa}} ‘spring’, not \textipa{\textit{*D-aRaT}}), while others avoid prefixing the initial feminine morpheme but preserve the suffixal marker (e.g., \textipa{\textit{RXeDm-eT}} ‘work’, not \textipa{\textit{*D-RXeDm-eT}}).

Building on our morphological account of countability --- where gender markers, within the morphological system, function as affixes to the countability markers ---, we propose a parallel analysis for definiteness. This unified perspective, while not standard in linguistics, is supported by the morphological behaviour of these markers, particularly their susceptibility to morphophonological influence. As an illustration, the English definite article \textit{the} has the allomorphs \emph{ðiː} and \emph{ðə}. In Riffian, the morpheme \textit{ð} undergoes allophonic variation to \textit{t} in the environment of a voiceless consonant, as in \textit{t-sa} ‘kidney’ (for comparison, before a voiced consonant: \textit{ð-ma} ‘edge’).

Beyond the morphophonological analysis, it is notable that several of these languages also display the same partial asymmetrical gender encoding observed in Riffian. Specifically, this asymmetry is characterised by the absence of an initial gender marker and/or the non-appearance of gender suffixation in some nouns. For example, in Spanish, proper nouns like \textit{Alejandr-o/a} ‘Alexander/Alexandra’ lack the initial gender marker, rather than appearing as \textit{*l-o/a Alejandr-o/a}. Similarly, nouns like \textit{l-a flor} ‘flower’ do not undergo gender suffixation (i.e., \textit{*l-a flor-a}), even though gender is marked by the article. This cross-linguistic phenomenon in unrelated synthetic languages suggests the operation of extra-phonological and extra-semantic principles.

Firstly, its occurrence in very different phonological environments makes it impossible to assume a common phonological explanation. Secondly, semantics is also unlikely to be the cause of these morphological irregularities. These languages do not share semantic features that could account for these patterns. At the same time, semantics tends to differentiate languages, whereas what we observe brings different languages together around a similar morphological phenomenon. Consequently, only a rational, unified explanation rooted in morphology can be reasonably assumed. Hence, we must unify the analysis of these synthetic languages, whether or not they involve definiteness or asymmetrical encoding\footnote{This includes languages with either full asymmetrical encoding (e.g., Arabic encodes gender only at the final position) or partial asymmetrical encoding.}, under a single theoretical framework, as proposed.

With this new perspective, it becomes possible to analyse the morphological patterns of both definiteness-marking (e.g., English, French, Greek, etc.) and non-definiteness-marking (e.g., Riffian, Russian, Hindi, etc.) synthetic languages in a unified manner\footnote{The canonical form for languages with definiteness is: $\bigotimes =$ definiteness $+$ countability $+$ radical $+$ gender $+$ number.}. This is especially relevant when considering how gender shifts affect these languages similarly during word formation (e.g., in Italian: \textit{la pentola} ‘pot.F’ $↬$ \textit{il pentolame} ‘cookware.M’;  \textit{il prete} ‘priest.M’ $↬$ \textit{la pretaglia} ‘set of priests.F’ \citep{Ludovico2020}, see also Section \ref{model_CN} for further details on Riffian).

In order to contextualise our research, the subsequent section will delineate the general landscape of linguistic theories pertinent to our work, and then proceed to an examination of formal morphology and mathematical models. This review will only examine the underlying theoretical principles of various approaches related to our subject.

\subsection{The lexemes and cognition: A review}\label{review_cog}

As a preamble, Figure \ref{fig0} presents a schematic representation of the linguistic system to contextualise the application domain of the TBMC model. The main components of this system form a linear chain of derivations from semantics to syntax. Each component functions independently and the outputs of each component are the inputs of the next. Furthermore, each part of this system interacts with other computational tools, namely word and meaning formation. The domain of these builders of items is clearly delimited as illustrated in Figure \ref{fig0}. 

The overall architecture is divided into two broad zones, distinguished by their position relative to the item-template pairing component: a pre-item-template pairing zone on the right, and a post-item-template pairing zone on the left. Each zone is enclosed within a dashed boundary and corresponds to a distinct stage of the linguistic derivation.

The pre-item-template pairing zone encompasses the earliest stages of the derivation. It contains two components: Semantics, represented as a diamond-shaped node to reflect its conceptual and non-computational nature, and Items (defined as form-meaning pairs or meaning sets), represented as a rounded rectangle. Together, these two components fall within the domain of Meaning formation. This zone is thus responsible for the construction of basic lexical units from semantic representations.

\begin{figure}[!h]
  \renewcommand{\thefigure}{\arabic{figure}}
  \caption{The structure of the linguistic system from meanings to sentences}
  \vskip 0.3cm
  \includegraphics[trim=0in 1.4cm 0in 5mm, clip, width=\textwidth]{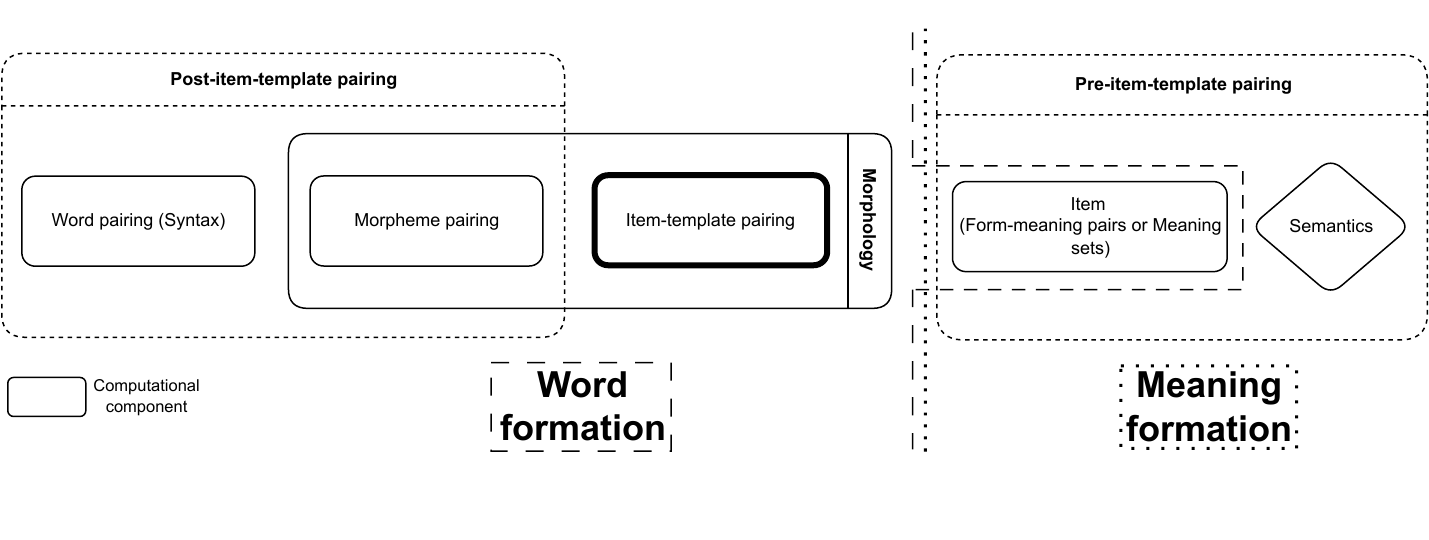}
  \label{fig0}
\end{figure}

The post-item-template pairing zone encompasses the later stages of the derivation and contains two computational components, each represented as a rounded rectangle: Morpheme pairing and Word pairing. The Item-template pairing component is visually distinguished by a bold border, indicating that this cognitive module is the focus of our study. The Item-template pairing and Morpheme pairing components together fall within the domain of Morphology. In turn, these two morphological components, together with Word pairing and Item, fall within the broader domain of Word formation, delimited by a large dashed box. Word pairing (Syntax) sits at the leftmost position in this chain, representing the component of syntactic structure. The Item component can interact with both the domains of Word formation and Meaning formation, which are viewed here as possessing distinct properties.

The figure thus illustrates a right-to-left ordering of the derivation: from the construction of meanings and primitive lexical items, through morphological operations, and finally to syntactic word pairing.

A central theoretical contribution embodied in this architecture is the distinction between the Item-template pairing and Morpheme pairing modules. Many existing frameworks treat morphology primarily in terms of realizational rules, that is, mappings between abstract morphosyntactic features and their phonological exponents. In the present model, this function is implemented by the Morpheme pairing component. However, the model argues that realizational rules alone are insufficient to account for all morphological phenomena. In particular, they cannot explain the inheritance and reassignment of morphological templates that accompany derivational processes.

This function is instead handled by the dedicated Item-template pairing component. During derivation, a newly formed lexical item must be associated with the template appropriate to its resulting lexical category. This distinction is especially important in synthetic languages, where derivational morphology plays a major role in lexical expansion. In such languages, derivation often involves a change of lexical category, for example, when a noun is derived from a verbal base. Because verbs and nouns typically instantiate different grammatical templates, the derived noun cannot simply retain the template of its source verb. Rather, it must acquire a template corresponding to its new nominal status.

The reassignment of templates in such cases cannot be reduced to realizational rules and therefore requires and justifies a separate computational mechanism. The Item-template pairing component is postulated to perform this function by assigning items to the templates associated with their lexical categories.

That being said, this study will focus solely on the item-template pairing and its formal modelling. Since this computational module is a newly proposed cognitive apparatus designed to capture morphological phenomena, it is essential to situate this component within the linguistic system. The next section reviews various approaches concerning morphological theory.

\subsubsection{Item-template pairing and morpheme pairing}\label{item_template}

The item-template pairing component occupies a central position within this sequence of computational units. The latter component is introduced as a new computational module of language and its formalism is presented later (see section \ref{method}). While we posit that this component belongs to the sphere of morphology, its role is to serve as an intermediary module between the components of morpheme pairing and item.

The item-template pairing component is not an element existing or even acknowledged in other theories, nor have alternative mechanisms been proposed to address it. As there are no competing proposals, a comparative analysis cannot be conducted. This lack of existing models for addressing this linguistic component is what motivates the proposal of the TBMC model.

The latter is particularly adapted to capture grammatical template shifts, as will be demonstrated later. In addition, it can generate item-template pairings directly, which constitutes one of its principal strengths. More specifically, the TBMC model handles both the learning and prediction stages within a single framework, without relying on external models.

It is important to emphasise this point because most theoretical frameworks in morphology --- and, more broadly, in linguistics, with the exception of computational linguistics and natural language processing --- rely on deductively formulated rules rather than mechanisms learned through a training process. Moreover, they generally do not incorporate algebraic operations into their algorithms. By contrast, both learning and algebraic computation constitute the foundation of the TBMC model.

Although not produced by a computational tool, the notion of item-template pairs is, however, found in other work. Before introducing the notion of item-template pairs, we need to revisit key linguistic concepts to understand the distinctions between the different theories of morphology.

It is generally accepted \citep[p. 49]{spencer1991} that the formation of linguistic facts typically falls into two categories: Item and Arrangement and Item and Process. Actually, it was observed that contemporary theoretical frameworks combine elements of both approaches. Behind these labels hide simple operations. The Item and Arrangement refers to a juxtaposition operation. As for the Item and Process, it may designate different operations, but let us just mention the concatenation operation, which is the main process used in linguistics. Concatenation and juxtaposition can both be categorised as a combination process. The former will fuse or map binary units (or more) into a single unit (e.g., concatenation(item$_1$, item$_2$) = item$_1^{*}$-item$_2^{*}$), while the latter will create from inputs other disjoint outputs (e.g., juxtaposition(item$_1$, item$_2$) = item$_1^{*}$, item$_2^{*}$).

With that in mind, similar differences regarding these operations are also noticed in morphology. From this meta-theoretical perspective, Distributed Morphology \citep{halle_distributed_1993} and Modern Word and Paradigm \citep{blevins_word_2016}, which are the two main theoretical branches in morphology in terms of scientific production \citep{siddiqi2016morphological}, can be distinguished on this basis. While not directly expressed, we deduce that these proposals implicitly involve both operations. Both models are only concerned with morpheme pairing; other dimensions of the linguistic system are not treated. However, their understanding of pairing of morphemes is not aligned.

Thus, it seems that each of them applies concatenation and juxtaposition on different linguistic facts. The items manipulated by their models are meanings and forms (viz., sequences of phonemes) and the two aim to assign phonological forms to meanings. In fact, they are the dual of each other. Thereby, Distributed Morphology employs concatenation on meanings and juxtaposes the forms. Conversely, in Modern Word and Paradigm, the meanings are juxtaposed and the forms undergo concatenation\footnote{To illustrate this meta-theory, consider the following straightforward example. In Distributed Morphology, concatenation($\sqrt{CAT}$, PL) = $\sqrt{CAT}$-PL; then juxtaposition(($\sqrt{CAT}$, /cat/), (PL, /s/)) = cat, s. While, in Modern Word and Paradigm, juxtaposition($\sqrt{CAT}$, PL) = $\sqrt{CAT}$, PL; then concatenation(($\sqrt{CAT}$, /cat/), (PL, /s/)) = cat-s. We must specify that, between these two operations, retrieval-like functions, specific to each framework, are used to obtain the phonological forms of the meanings. Finally, when spoken, both linguistic structures are pronounced continuously. However, according to these theories, the concatenative structure constitutes a single unit, while the other does not reflect this cognitive distinction.}. Even though not explicitly stated in this way in their different writings, our analysis suggests that the primary distinctions between these approaches occur at this level.

No theory either in syntax or morphology has offered an explanation for the processes of selecting and pairing grammatical meanings\footnote{The set of grammatical meanings should be understood in contrast to lexical meanings. Grammatical meanings are realised by modifying the lexical form (i.e., the root) in various ways, such as affixation or phonological changes. In Generative Grammar, they are divided into several categories: categorial, case, and φ-features. This set is language-specific; in the case of the Riffian language and similar languages, categorial and φ-feature elements are included. However, from a theoretical and formal standpoint, we do not consider this set limited to these semantic features, and prefer to leave it open to other types of grammatical features. The definition of set $G$ (see Supplementary Appendix A) reflects this openness.} to items to build item-template pairs. Nor have rational justifications been provided for the dynamic mapping of items onto grammatical templates engendered during word and meaning formation. Consequently, as said, all existing theories lack a computational mechanism for generating item-template pairs. As the reader may understand it now, this computing layer should not be seen as opposed to the morpheme pairing but more as a complementary module.

While these approaches do not feature such a component, the notions of template and item-template pairs are prevalent in linguistic theory. Such a linguistic fact is often used by various theories, for example, the concept of Complex Symbol (\citeyear{chomsky_aspects_1965}) or Lexical array (\citeyear{chomsky_minimalist_1995}), proposed by Chomsky. Other researchers have proposed similar concepts under different names, such as Morphosyntactic Representation \citep{anderson_-morphous_1992}, Morphosyntactic array \citep{aronoff_morphology_1993}. Despite the variation in terminology, these concepts share very similar attributes.

We would like to finish this review by mentioning work in formal language theory and artificial intelligence.

\subsubsection{Formalism and Functions with set}\label{sets}

The mathematical formalisation of morphology is not a new idea, with Aronoff (\citeyear{aronoff_morphology_1993}) being one of the first to propose the use of functions. However, these functions do not establish a relationship between a set of items and a set of templates. Instead, following the Separation Hypothesis  \citep{beard_lexeme-morpheme_1995}, these approaches focus solely on the relationship between the form of a lexeme and its affixes. Similarly, we acknowledge that the morphology component includes a generative mechanism. In other words, it comprises a computative lexicon that not only deals with morpheme pairing but also enables item-template pairing.

The primary criticism we can raise against these models lies in their incomplete formalisation. This gap probably arises from the apparent mismatch between the non-linearity of data and the use of ordinary functions, which may not fully capture this complexity. While numerous mathematical proposals exist in linguistics, the majority concentrate on syntax. In addition, these works are generally grounded in automata theory. For instance, several tree-automata-based models \citep{stabler1997,graf2011} have emerged within the Minimalist Program framework, which includes Distributed Morphology. This approach departs significantly from the foundational assumptions of our model. The main difference is that the elements involved in the mapping are sets. Such a map is named set-valued set function (or set to set function) in the literature.

The topic of set-valued set functions has garnered considerable attention and led to significant contributions in various scientific fields, including control systems, mathematical economics, and game theory, among others \citep{Obukhovskii_2020}. The similarities between the TBMC model and these approaches are merely limited to the use of this type of function. In our case, the set elements are not real values, but rather symbolic information. Additionally, similar to Deep Learning \citep{Ovidiu_2020}, our model is a function composition, which is needed to overcome the non-linear mapping between the item and the template.

Even though they are not directly related to topics studied in computational and theoretical linguistics, we would like to mention works published in the field of Artificial Intelligence, particularly in Natural Language Processing, that have also explored the use of sets as function arguments. Thus, recent advances in Artificial Neural Network have utilised sets, rather than vectors, to learn the parameters of a model \citep{Vaswani_2017,Zaheer2017}. Beyond this specific point, the mathematical approach used in this article differs considerably from traditional Deep Learning methods. In contrast, the TBMC model uses sets as both input and output, without involving optimisation-based learning processes.

Therefore, the TBMC model is an ideal candidate for modelling item-template pairing. Unlike conventional mathematical models, it processes symbolic information in its discrete form, obviating the need for conversion to continuous values. Moreover, it can handle sets of symbolic information (e.g., templates of meanings) and apply set-theoretic operations to compute outputs (e.g., item-template pairing), which other mathematical models cannot achieve. Its ability to handle non-linearity without requiring extensive learning is particularly valuable, as it does not need large quantities of paired data and labels. This feature is especially beneficial for low-resource languages and makes the underlying hidden information used for prediction explainable.

The review has helped us to establish the originality of the TBMC model and its application to overlooked linguistic issues. Likewise, this broad perspective allows us to identify connections between the TBMC model and other approaches, suggesting potential applications of set-valued set functions in other fields.

\section{Formal analysis and Results}\label{data}

This section explores the relationship between nouns and gender in Riffian. Dynamical grammatical gender shift has not been studied in Riffian or Berber linguistics. This analysis offers a novel exploration of the topic. Our analysis focuses on nouns of action (NA), countable nouns (C), and uncountable nouns (U)\footnote{Other terminologies are used to name these classes of nouns, namely: count noun for countable noun and non-count noun for uncountable noun.}. To provide a comprehensive analysis, we have included numerous examples. While some might find this excessive detail, this is justified to prevent confirmation bias and ensure a thorough understanding of Riffian's rich morphological diversity. Data were collected from the Ayt Waryaghel variety (based on the author's native variety) and the Ayt Said variety (primarily from Serhoual's (\citeyear{serhoual_dictionnaire_2002}) dictionary). While Riffian lacks standardisation, this sample provides a representative overview of gender marking in Riffian. To model these data, we employ the TBMC model, introduced in the next section. This model is capable of accurately handling the multiple grammatical shifts (i.e., sets of semantic elements) that occur during both intra-lexical and inter-lexical derivation. For example, in the case where singulative and feminine markers appear after conversion: \textipa{\textit{t:ef:a\textcrh}}~‘apples’ $\rightarrow$ \textipa{\textit{Da-t:ef:a\textcrh T}}~‘an apple’  (see also the example (2) in \nameref{intro}).

\subsection{The TBMC model}\label{method}

A new mathematical formalism is used to capture the phenomenon of grammatical gender shift. The purpose of this model is to pair items onto grammatical templates when items undergo a word and/or meaning formation. This formalism is not solely based on a set-valued set function; it also incorporates other characteristics (see Supplementary Appendix A). Thus, the TBMC model is a six-tuple $\mathbb{M}$ $=$ $\langle W, K, T, f, g, h\rangle$ where:

\begin{itemize}

\item $W$ is the set of items, formally defined as a set of sets of symbols; $K$ is the set of the retrospective determinants of shifting templates, formally defined as a set of tuples of symbols; and $T$ is the set of grammatical templates, formally defined as a set of sets of symbols.

\item $f$ is the backward recursive function. It is a tuple-valued set function such that $f: W \rightarrow K$   

\item $g$ is the gradient function. It is a set-valued multiary function such that $g: K \rightarrow T$

\item $h$ is the morphosyntactic transfer function such that  $h = g \circ f$. It is a set-valued set function such that $h: W \rightarrow  T$

\end{itemize}

In addition, all item-template pairs must satisfy specific gradient conditions (see Definition~\ref{def2}). These parametric conditional factors are related to the sets $K$ and $T$, and particularly the subset of meanings $M$. Likewise, the items of $W$ are constituted by elements from $M$. In other words, for $w \in W, \exists\{m_i\}_{i \in \mathbb{N}} \subset M, \{m_i\}_{i \in \mathbb{N}} \subset w$, and for $t \in T, \exists\{m_j\}_{j \in \mathbb{N}} \subset M, \{m_j\}_{j \in \mathbb{N}} \subset |t|$. In general, the gradient condition involves set operations between $K$ and $T$. Those operations allow us to compute $t \in T$ for a given $w \in W$. It can be defined as follows:

\begin{deflist}
\addtocounter{deflist}{3}
 \refstepcounter{deflist} \label{def2} 
 
   \justifying

    \item[Definition 4]\hfill

    Let us consider this definition of the universal gradient condition algorithm:

    \begin{itemize}[labelsep=5mm,parsep=0.3mm,itemsep=2mm]

\item[\(\diamond\)] Let $C_g = \{\mathit{gradcond}_{W}^{n} (f,h; \op, w) \;|\; w \in W, \op \in \mathcal{O}_{\mathrm{alg}}\}$ be the set of generated set parameters. Here, $\mathit{gradcond}_{W}^{n} (f,h; \op, w) = p^{n}$ is a binary function mapping to the power set $\mathcal{P}(\Gamma)$, where:

\begin{itemize}[labelsep=5mm,parsep=0.3mm,itemsep=2mm]

\item[\(\centerdot\)] $\mathcal{O}$ is the set of all admissible operator functions $\op(\cdot,\cdot)$. The function $\op(\cdot,\cdot)$ denotes an algebraic operation ($\mathcal{O_{\mathrm{alg}}} \subset \mathcal{O}$), a logical expression, or a choice function. In the present case, for the determination of grammatical templates, $\op(\cdot,\cdot)$ corresponds to the symmetric difference operation, symbolised by $\Delta$. The properties below assume $\op$ is algebraic (cf. $\mathcal{O_{\mathrm{alg}}}$).

 \item[\(\centerdot\)] For any $w \in W$ with $f(w) \in K$, the element $h(w) = t_j$, where $t_j \in T$, must satisfy $g(f(w)) = t_j$ $~\Leftrightarrow~$ $\op(f(w), t_j)$ $= p^{n}$, where $p^{n}  \in \mathcal{P}(\Gamma)$.
 
 \item[\(\centerdot\)] Depending on the specific form of $\op(\cdot,\cdot)$, the gradient condition is either satisfied by updating iteratively $t_j \in T$ until the equality holds, or by directly solving the equation defined by $\op(\cdot,\cdot)$. 

\item[\(\centerdot\)] The superscript $n$ is an indexing value assigned to $\mathit{gradcond}$ and the set parameter $p$, such that $n, m \in \mathbb{N}$ with $n \neq m \implies p^{n} \neq p^{m}$. Let us note that $\mathit{gradcond}_{W}^{n}$ and $p^{n}$ have the same indexing. Hence, the choice of one specific $\mathit{gradcond}_{W}^{n}$ and $p^{n}$ (or more) depends on the elements of $K$ and $T$.

\end{itemize}

\end{itemize}

\end{deflist}

Those gradient conditions may vary across languages, however, some of them are common to multiple languages (see \nameref{intro}). Thus, the universal canonical form of the gradient condition for grammatical gender shift is presented in Section \ref{shift} (see Gradient Condition \ref{gradcond2}). Similarly, the gradient condition for semantic widening, which results in non-gender shift, is also provided in the same section (see Gradient Condition \ref{gradcond3}).

In what follows, we establish the existence and uniqueness of the gradient conditions by invoking the algebraic properties of the symmetric difference ---namely, \emph{associativity}, \emph{commutativity}, \emph{inverse}, and \emph{identity} \citep[see][Chapter~2, Proposition~2.3, p.~10]{GilbertNicholson2004} together with the structural properties of the TBMC model (see Definition~1-3 in Supplementary Appendix~A). We begin with the proof of existence for $p$. 

\renewcommand{\thethmd}{\arabic{thmd}}  
\begin{thmd} Let  $T \subset \mathcal{P}(M)$ ~and~ $\mathcal{P}(\Gamma) \subset \mathcal{P}(M)$, If $t_j$ and $t_i$ are sets, $t_j$ and $t_i$ $∈ T$, then there exists a set $p$ $\in$ $\mathcal{P}(\Gamma)$, such that $t_i$ $\Delta$ $t_j$ $= p$. \end{thmd} 

\begin{proof} 

In set theory, the symmetric difference of two sets is not a primitive operation; it can be defined using the basic operations of  \emph{union}, \emph{intersection} and \emph{difference}. Consequently, the symmetric difference $t_i$ $\Delta$ $t_j$ admits (at least) two equivalent formulations in terms of these operations:
	
\begin{align*}
\ddag: & \,\, (t_i \setminus t_j)  \cup (t_j \setminus t_i) \\
\ddag\ddag: & \,\, (t_i \cup t_j) \setminus (t_i \cap t_j)\\ 
\end{align*}

Since $t_i$, $t_j$ $\in \mathcal{P}(M)$, we have $t_i$ $\subset M$ and $t_j$ $\subset M$. Therefore, any of the standard set operations applied to $t_i$ and $t_j$ yields a set that is still a subset of $M$, hence also an element of $\mathcal{P}(M)$. It follows that the power set $\mathcal{P}(M)$ is closed under those operations. Thus, $p$ exists as an element of $\mathcal{P}(M)$ by the axioms of Set Theory \citep{Suppes1972}. To prove its existence, we demonstrate this constructively using, first, definition (\ddag): 

\leqnomode
\begin{flalign*}
&\tag{i} \hskip 3mm t_i \setminus t_j = \{m  \mid m  \in t_i \text{ and } m \notin t_j\} \text{ exists by the axiom schema of separation, since it derives from set $t_i$.} \\
&\tag{ii} \hskip 3mm \text{Similarly for } t_j \setminus t_i \text{ (see above (i)).} \\
&\tag{iii} \hskip 3mm (t_i \setminus t_j)  \cup (t_j \setminus t_i)  =   \{m  \mid m  \in t_i \setminus t_j  \text{ or } m \in t_j \setminus t_i \} \text{ exists by the axiom of union.} 
\end{flalign*}

Then, according to (i)-(iii), $p$ exists as a set. The same constructive proof can be carried out using the second equivalent definition (\ddag\ddag). 

\begin{flalign*}
\tag{l} & \hskip 2mm t_i\, \cup\, t_j =  \{m  \mid m  \in t_i \text{ or } m \in t_j \} \text{ exists by the axiom of union.} \\
\tag{ll} & \hskip 2mm t_i\, \cap\, t_j =  \{m  \mid m  \in t_i \text{ and } m \in t_j \} \text{ exists by the axiom schema of separation.} \\
\tag{lll} & \hskip 2mm  (t_i\, \cup\, t_j)\,  \setminus\,  (t_i\, \cap\, t_j)  =   \{m  \mid m  \in t_i\,  \cup\, t_j  \text{ and } m \notin t_i\,  \cap\, t_j \} \text{ exists by the axiom schema of separation.} 
\end{flalign*}

\reqnomode 

Then, in either case, the existence of $p$ follows immediately from the closure of $\mathcal{P}(M)$ under union, intersection, and difference.

\end{proof}

Having established existence, it remains to show that no other element satisfies the same defining equation by proving that any two elements satisfying it must coincide, thereby ensuring uniqueness.

\begin{thmd} Let $t_j$ and $t_i$ $∈ T$ with $t_j\setminus \{N\} $ and $t_i\setminus \{N\} $ $\in$ $\mathcal{P}(\Gamma)$, then any  $p$ $\in$ $\mathcal{P}(\Gamma)$ satisfying $t_i$ $\Delta$ $t_j$ $= p$ is unique. \end{thmd} 

\begin{proof} Suppose, for contradiction, that the value of \( t_i \Delta\ t_j \) is not unique. Then there exist sets \(p_1\) and \(p_2\) with \( p_1 \neq p_2\) such that \( t_i \Delta\ t_j = p_1\) (\(\dagger\)) and \( t_i \Delta\ t_j = p_2\) (\(\dagger\dagger\)).

(a) As will be shown in Lemma (1) \emph{infra}, for the precise formulation, the equation (\(\dagger\)) can be rewritten as: \( t_i \Delta\ p_1 =  t_j\)

(b) We put equation (\(\dagger\)) into equation (\(\dagger\dagger\)) and we argue as follows:

\begin{align*}
t_i \Delta\ t_j & = p_2 \\
t_i \,\Delta\, (t_i \,\Delta\, p_1) &= p_2 && \text{(Substitute } t_j \text{ by } t_i \,\Delta\, p_1 \text{, see (a))}  \\
(t_i \,\Delta\, t_i) \,\Delta\, p_1 &= p_2 && \text{(Associativity)} \\
\emptyset \,\Delta\, p_1&=p_2 && \text{(Inverse: } t_i \,\Delta\, t_i = \emptyset) \\
p_1 &= p_2  && \text{(Identity: } \emptyset \,\Delta\, p_1= p_1)
\end{align*}

Thus, this shows \( p_1 = p_2\), which contradicts the assumption that \( p_1 \neq p_2\). Therefore, the value of \( t_i \Delta\ t_j \) is uniquely determined.

\end{proof}

Both theorems serve as a universal guarantee that, for any given grammatical shift, a unique gradient operand $p$ exists that maps items onto their respective templates. This demonstrates the expressive power of our model and its capacity to represent diverse grammatical shifts effectively. This deductive approach is conceptually analogous to the Universal Approximation Theorem for neural networks, which mathematically guarantees that a given architecture can approximate any continuous function under specified conditions \citep{cybenko1989}. In both cases, the key property is representational completeness (or capacity) of a parametric family, manifested in our framework through the existence (and uniqueness) of $p$, and in neural networks through the density of function approximators.

Furthermore, alongside this model, we have established a prototypical process chain of template shifts. Assuming the starting conditions, each chain initiates at an input head and proceeds through successive directional derivatives along a specific grammatical element. A rooted phylotemplatic tree is used to illustrate these derivations (see Figure \ref{fig_new}). Directional derivatives arise from the processes involved in the formation of words and meanings. The latter are four in total, specifically: conversion, borrowing, morphological derivation\footnote{We propose a distinction between morphological derivation, involving changes to the radical form of a word, and conversion, where the radical form remains unchanged. These modifications include sound changes to the root or its affixes, and alterations to the morphological structure (e.g., scheme), excluding grammatical markers like those for feminine or plural forms.} and semantic widening. Each is, respectively, symbolised by function symbols: ${\to}(\cdot)$, ${\Lsh}(\cdot)$, ${\looparrowright}(\cdot)$ and ${\supseteq}(\cdot)$\footnote{Although the semantic widening function and the subset operator use the same symbol $⊇$, they differ in both their properties and notation. The subset operator is binary and uses infix notation, whereas the semantic widening function is n-ary and uses prefix notation.}. They must be viewed as n-ary function symbols (e.g., $\rightarrow\hspace{-1.4mm}(x_1, ..., x_n)$).

The primary property of these functions is to clone elements of item-template pairs to create new items. Besides this common characteristic, their purposes are not identical. Thus, different linguistic facts may result from these functions depending on word and meaning formation employed. Since this article focuses mainly on semantic widening and conversion, we will restrict ourselves to these processes to clarify their objectives. In the case of semantic widening, the function modifies the base item without altering the cardinality of $W$. While with conversion, the features of the base item are preserved, and both the resultant and base items become elements of $W$. Formally, this can be expressed as follows:

\begin{deflist}
 \addtocounter{deflist}{-1}
  \refstepcounter{deflist} \label{pro2}
  
  \justifying

    \item[Proposition 2] \hfill
    
   Let us propose a formal definition for conversion and semantic widening:

\begin{alphalist}[labelsep=2mm,topsep=0pt,parsep=0.3mm,itemsep=5mm]

\item Conversion: If there exist $\text{item}_i \in W$, $\#W=n$ and ${\rightarrow}(\text{item}_i) = \text{item}_j$, then it follows that $\text{item}_i \neq \text{item}_j$, $\text{item}_i \text{ \& } \text{item}_j \in W$ and $\#W=m$, $m=n+1$ with $\forall n \in \mathbb{N}$ 

\item Semantic widening: If there exist $\text{item}_i \in W$, $\#W=n$ and ${\supseteq}(\text{item}_i) = \text{item}_j$, then it follows that $\text{item}_i \supseteq \text{item}_j$, $\text{item}_i \notin W \text{ \& } \text{item}_j \in W$ and $\#W=m$, $m=n$ with $\forall n \in \mathbb{N}$

 \end{alphalist}

\end{deflist}

\subsection{Initial templates and methodological considerations}\label{init_temp}

A strong assumption of our theory is that template shifts are deterministic. In other words, by considering the properties of the base word (including both the item and template), the properties of the resultant word (item only), and the formation process that links them, we can accurately determine the template output for the resultant word. This determinism is a significant contribution of our model, as it enables not only the computation of successive derivatives but, more importantly, also the successive antiderivatives. Similar to other computational models, the results of iterative derivations depend on the initial inputs, also referred to as initial values or conditions. Hence, further intra-lexical derivations are contingent on these initial conditions and must be determined.

To achieve this, it is essential, for each syntactic category and cognitive set (see Proposition 1 in Supplementary Appendix A), to identify the initial templates involved in inter-lexical derivations from the input heads. Given that no computational methods currently exist for this task, we decided to develop a new approach based on empirical observation and quantitative analysis. Specifically, we want to identify the most frequent grammatical template for each cognitive set. Since our focus is on grammatical gender shift, we will limit our analysis to verb-to-noun derivations. Furthermore, as language evolves over time through various derivations, considerable morphological discrepancies arise. This, in turn, makes it difficult to ascertain these parameters solely through frequency analysis. Selecting the appropriate templates to serve as initial conditions depends on numerous factors, so specific methodological criteria have been established.

This novel heuristic approach employs dimension reduction \citep{jia_feature_2022} based on both domain knowledge and empirical data \citep{Mao2010}. The goal is to reduce the feature dimensions of items by leveraging both linguistic and extra-linguistic knowledge\footnote{We recommend following this methodology, as it can be generalised to any synthetic language. However, the number and type of reduced features can be adapted to the language under study. Moreover, this method is not a prerequisite for applying the TBMC model; other approaches can also be considered to determine the initial templates.}. This feature selection results in a subspace of items with low dispersion and feasible initial solutions. Such a dimension reduction method is conceptually similar to techniques used in various fields of Applied Mathematics. For example, in Machine Learning, centroid initialisation for the K-means algorithm can be informed by Principal Component Analysis \citep{Xu2015}. Accordingly, we aim to determine the number of initial templates and their semantic features. For the Riffian language, the total solution space consists of $2^{(\#{\scriptscriptstyle\bigotimes} - 1)}$ templates (i.e., 64 templates). The objective is to select the initial templates from these candidates.

\begin{figure}[htbp]
  \renewcommand{\thefigure}{\arabic{figure}}
   \caption{Conceptual illustration of the proposed subgroup optimisation methodology.}
    \centering
    \includegraphics[width=\textwidth]{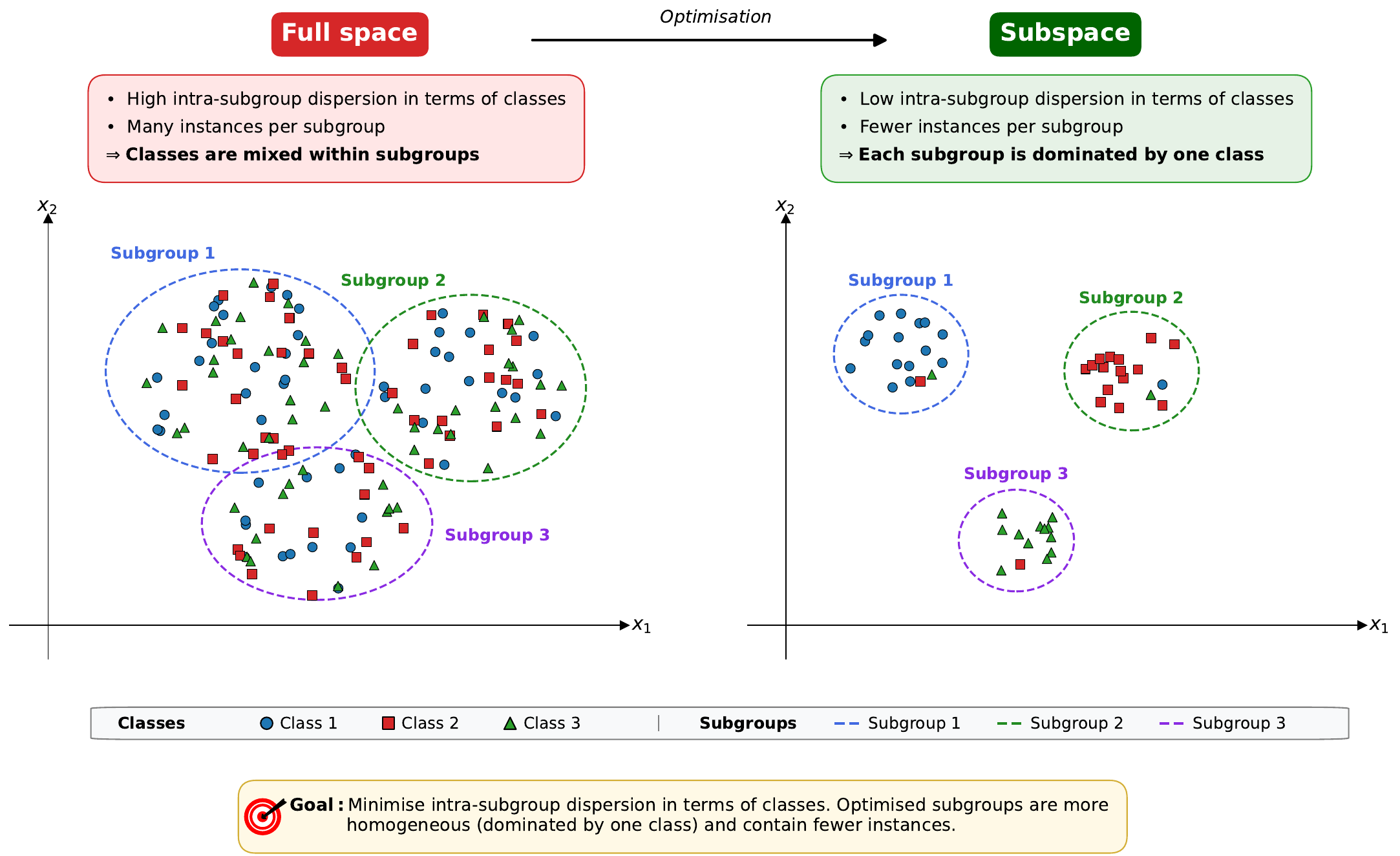}
   
    \label{fig:cluster_optimization}
\end{figure}

Overall, the approach is driven by an optimisation objective that seeks to minimise intra-subgroup dispersion. Figure \ref{fig:cluster_optimization} provides a schematic illustration of this goal. In other words, the aim is for each subgroup to be dominated by a single class, with instances from other classes appearing only marginally, an objective closely related to subgroup discovery \citep{Helal2016}\footnote{The reference to these machine learning methods is intended to help the reader understand the methodological concepts underlying the proposed approach. We do not claim to use the same algorithms or to produce the same outputs; rather, the comparison is intended to highlight conceptual similarities.}. Under this assumption, we can determine the initial template associated with a subgroup by assigning it the dominant class within that subgroup. The target classes correspond to grammatical templates, while the instances themselves are deverbal items. Because these items are grouped into cognitive sets, the prediction task can be formalised as follows:

\begin{equation}
\textit{InitialTemplate} : \{C, U, NA\} \rightarrow T,
\end{equation}

where $C$, $U$, and $NA$ denote the cognitive sets \textit{Countable Noun}, \textit{Uncountable Noun}, and \textit{Noun of Action}, respectively, and $T$ denotes the set of grammatical templates.

\subsubsection{Data exploration and analysis}\label{data_exploration}

Prior to applying the proposed technique, we first examine the relationship between nouns and templates. To this end, a dataset was compiled. This section presents the dataset, outlines its main characteristics (see Table~\ref{tab:pivot}), and provides a quantitative analysis that motivates the use of our methodology. The dataset comprises attested Riffian nouns associated with several linguistic and extra-linguistic features (see Supplementary Appendix B for details and Table \ref{tab:variables}). For the purposes of the present analysis, only the cognitive set and template variables are considered.

\begin{table}[ht!]
\centering
\caption{Semantic features by grammatical templates}
\label{tab:pivot}
\begin{tabular}{llrrrrrrrr r}
\hline\hline
 & & \multicolumn{9}{c}{\textbf{Grammatical templates}} \\
\cline{3-11}
 & \textbf{Semantic features} & \textbf{$T_1$} & \textbf{$T_2$} & \textbf{$T_3$} & \textbf{$T_4$} & \textbf{$T_5$} & \textbf{$T_6$} & \textbf{$T_8$} & \textbf{$T_9$} & \textbf{Total} \\
\hline\hline
 & Countable                      &  81 & 52 &  0 &  0 & 14 & 17 & 0 &  0 & 164 \\
 & Countable/Uncountable          &   0 &  3 &  0 &  1 &  2 &  8 & 0 &  0 &  14 \\
 & Noun of action                 & 150 &  3 &  0 &  1 &  3 &  3 & 0 &  1 & 161 \\
 & Noun of action/Countable       &   0 &  2 &  0 &  0 &  0 &  1 & 0 &  0 &   3 \\
 & Noun of action/Uncountable     &   4 &  1 &  0 &  1 &  1 &  0 & 0 &  0 &   7 \\
 & Uncountable                    &   8 & 16 & 11 & 17 &  5 & 40 & 2 & 11 & 110 \\
\hline\hline
 & \textbf{Grand total}           & \textbf{243} & \textbf{77} & \textbf{11} & \textbf{20} & \textbf{25} & \textbf{69} & \textbf{2} & \textbf{12} & \textbf{459} \\
\hline\hline
\end{tabular}
\end{table}

Table~\ref{tab:pivot} presents the distribution of 459 items across six semantic features and eight grammatical templates ($T_1$–$T_9$, with $T_7$ absent from the data). The corpus is dominated by two cognitive sets: Countable ($N = 164$, $35.7\%$) and Noun of action ($N = 161$, $35.1\%$). They together account for over $70\%$ of all observations, while hybrid categories (Countable/Uncountable, Noun of action/Countable, Noun of action/Uncountable) remain marginal, collectively representing fewer than $6\%$ of the total. Across grammatical templates, $T_1$ is by far the most frequent ($N = 243$, $52.9\%$), followed distantly by $T_2$ ($N = 77$, $16.8\%$) and $T_6$ ($N = 69$, $15.0\%$), whereas templates $T_3$, $T_4$, $T_8$, and $T_9$ exhibit notably low frequencies, suggesting their limited productivity in the sampled corpus. The uncountable subgroup constitutes the third largest group ($N = 110$, $23.9\%$) and exhibits a notably more dispersed profile, with non-negligible counts across $T_2$, $T_4$, $T_6$, and $T_9$, pointing to greater grammatical versatility for uncountable nouns.

A clear interaction between cognitive set and template is observable. We can notice that countable nouns cluster predominantly in $T_1$ and $T_2$, nouns of action are strongly concentrated in $T_1$ (150 out of 161 items), and uncountable nouns, while also most frequent in $T_1$, show a comparatively broader distribution across templates, most notably $T_6$ ($N = 40$), $T_4$ ($N = 17$), $T_2$ ($N = 16$), and $T_9$ ($N = 11$), indicating greater grammatical flexibility for this category. These patterns suggest that cognitive nominal classification is a significant predictor of grammatical template selection.

\begin{table}[H]
\centering
\caption{Chi-square goodness-of-fit results per cognitive set (H\textsubscript{0}: uniform distribution across templates).
$k$ = number of non-zero templates; $N$ = total item count.
Significance codes: $^{***}p < .001$; $^{*}p < .05$; \textit{ns} $p \geq .05$.}
\label{tab:gof}
\begin{tabular}{lrrrrl}
\hline\hline
\textbf{Semantic features} & $N$ & $k$ & $\chi^2$ & $df$ & $p$ \\
\hline\hline
Countable                   & 164 & 4 &  73.805 & 3 & $< .001^{***}$ \\
Noun of action              & 161 & 6 & 678.59 & 5 & $< .001^{***}$ \\
Uncountable                 & 110 & 8 &  70.364 & 7 & $< .001^{***}$ \\
\hline
Countable/Uncountable       &  14 & 4 &   8.286 & 3 & $.04^{*}$     \\
Noun of action/Countable    &   3 & 2 &   0.333 & 1 & $.564$ \textit{ns} \\
Noun of action/Uncountable  &   7 & 4 &   3.857 & 3 & $.277$ \textit{ns} \\
\hline\hline
\end{tabular}
\end{table}

To assess whether the distribution of grammatical templates within each cognitive set departs from uniformity, we conducted a series of chi-square goodness-of-fit tests, one per noun type, treating the observed counts across non-zero templates as the test distribution and an equal-frequency allocation as the null hypothesis (H\textsubscript{0}).

The goodness-of-fit results reported in Table~\ref{tab:gof} reveal a clear divide between the cognitive sets in terms of both statistical reliability and theoretical relevance. The three cognitive sets that emerge as statistically robust, countable ($\chi^2(3) = 73.81$, $p < .001$), noun of action ($\chi^2(5) = 678.59$, $p < .001$), and uncountable ($\chi^2(7) = 70.36$, $p < .001$), are also the three most frequent categories in the corpus. Their frequencies are sufficiently large to afford reliable distributional estimates, and their rejection of the uniformity hypothesis is unambiguous. During verb-to-noun derivation, which is the focus of this analysis, deverbal nouns can only belong to these cognitive sets, hence they are of great importance in the subsequent part of the study.

The remaining three semantic features, countable/uncountable ($\chi^2(3) = 8.29$, $p = .04$), noun of action/ countable ($\chi^2(1) = 0.33$, $p = .564$), and noun of action/uncountable ($\chi^2(3) = 3.86$, $p = .277$), are collectively characterised by marginal or absent item counts. The low frequency in the present corpus may itself reflect their peripheral status in the mental lexicon. Moreover, noun of action/countable and noun of action/uncountable did not reach significance. With sample sizes ranging from 3 to 14 items, these subgroups are too sparsely attested for the chi-square approximation to be trustworthy. The expected cell frequencies fall well below the conventional threshold of 5, inflating the risk of both Type I and Type II errors \citep{Cochran1952}. Retaining them in subsequent analyses would therefore risk distorting the distributional patterns observed for the prototypical cognitive sets without adding interpretable evidence.

Beyond the purely statistical concern, these hybrid categories (i.e., countable/uncountable, noun of action~/ countable, and noun of action/uncountable) do not constitute canonical cognitive sets. These items undergo semantic widening from a noun base. Since these hybrid categories does not occur between a base verb and a resultant noun, they fall outside the scope of our analysis. Instead, this section focuses exclusively on deverbal nouns to determine their initial templates. For these reasons, the following sections restrict the analysis to the three prototypical cognitive sets (i.e., countable, noun of action, and uncountable), which together account for $435$ of the $459$ items ($94.8\%$) in the dataset, and which are the only sets considered in the post-hoc analysis.

For the three noun types where uniformity was rejected, we performed pairwise two-proportion $z$-tests on all possible template pairs, applying Bonferroni correction to control the family-wise error rate (see Table \ref{tab:bonferroni}). For countable nouns, $T_1$ ($\hat{p} = 0.494$) was significantly more frequent than other grammatical templates. For nouns of action, $T_1$ was also overwhelmingly dominant ($\hat{p} = 0.932$), differing significantly from every other attested template ($z \geq 16.4$, all $p < .001$), whereas no pair among the remaining five templates reached significance. However, it is worth noting that $T_2$ is also significantly above $T_5$ and $T_6$. For uncountable nouns, $T_6$ ($\hat{p} = 0.364$) stood out as the strongly preferred template, differing significantly from other grammatical templates. Taken together, these findings indicate that non-uniform template preferences are robust and replicable across the three cognitive sets, with $T_1$ functioning as the dominant template for countable nouns and nouns of action, and $T_6$ playing an analogous role for uncountable nouns.


\begin{table}[H]
\centering
\caption{Significant pairwise two-proportion $z$-tests with Bonferroni correction for the three prototypical cognitive sets. Only pairs that survive correction ($p < \alpha_{\text{adj}}$) are reported; all remaining pairs were non-significant. $\hat{p}$ = observed proportion within the cognitive set; $\alpha_{\text{adj}} = 0.05/k$, where $k$ is the total number of pairs per set. Full results are provided in the Supplementary Appendix C1.}
\label{tab:bonferroni}
\begin{tabular}{llrrrrl}
\hline\hline
\textbf{Cognitive set} & \textbf{Pair} & $\hat{p}_i$ & $\hat{p}_j$ & $z$ & $p$ & \\
\hline\hline
\multicolumn{7}{l}{\textit{Countable} ($N = 164$,\; $k = 6$,\; $\alpha_{\text{adj}} = 0.0083$)} \\
\hline
 & $T_1$ vs $T_2$ & 0.494 & 0.317 &  3.261 & $.001$    & $^{**}$  \\
 & $T_1$ vs $T_5$ & 0.494 & 0.085 &  8.156 & $< .001$  & $^{***}$ \\
 & $T_1$ vs $T_6$ & 0.494 & 0.104 &  7.72   & $< .001$  & $^{***}$ \\
 & $T_2$ vs $T_5$ & 0.317 & 0.085 &  5.234 & $< .001$  & $^{***}$ \\
 & $T_2$ vs $T_6$ & 0.317 & 0.104 &  4.742 & $< .001$  & $^{***}$ \\
\hline
\multicolumn{7}{l}{\textit{Noun of action} ($N = 161$,\; $k = 15$,\; $\alpha_{\text{adj}} = 0.0033$)} \\
\hline
 & $T_1$ vs $T_2$ & 0.932 & 0.019 & 16.404 & $< .001$  & $^{***}$ \\
 & $T_1$ vs $T_4$ & 0.932 & 0.006 & 16.639 & $< .001$  & $^{***}$ \\
 & $T_1$ vs $T_5$ & 0.932 & 0.019 & 16.404 & $< .001$  & $^{***}$ \\
 & $T_1$ vs $T_6$ & 0.932 & 0.019 & 16.404 & $< .001$  & $^{***}$ \\
 & $T_1$ vs $T_9$ & 0.932 & 0.006 & 16.639 & $< .001$  & $^{***}$ \\
\hline
\multicolumn{7}{l}{\textit{Uncountable} ($N = 110$,\; $k = 28$,\; $\alpha_{\text{adj}} = 0.0018$)} \\
\hline
 & $T_1$ vs $T_6$ & 0.073 & 0.364 & $-$5.224 & $< .001$ & $^{***}$ \\
 & $T_2$ vs $T_6$ & 0.145 & 0.364 & $-$3.715 & $< .001$ & $^{***}$ \\
 & $T_3$ vs $T_6$ & 0.1     & 0.364 & $-$4.633 & $< .001$ & $^{***}$ \\
 & $T_4$ vs $T_6$ & 0.155 & 0.364 & $-$3.539 & $< .001$ & $^{***}$ \\
 & $T_5$ vs $T_6$ & 0.045 & 0.364 & $-$5.85 & $< .001$ & $^{***}$ \\
 & $T_6$ vs $T_8$ & 0.364 & 0.018 &  6.519   & $< .001$ & $^{***}$ \\
 & $T_6$ vs $T_9$ & 0.364 & 0.1     &  4.633   & $< .001$ & $^{***}$ \\
\hline\hline
\multicolumn{7}{l}{\footnotesize $^{**}p < \alpha_{\text{adj}}$;\quad $^{***}p < .001$. Negative $z$ indicates $\hat{p}_i < \hat{p}_j$.}
\end{tabular}
\end{table}

So far, we have considered all nouns in the sample regardless of their morphological status, that is, without distinguishing between deverbal and non-deverbal formations. We now want to conduct the same analysis restricted to deverbal nouns in order to determine whether they pattern in the same way as the full sample. The goodness-of-fit tests confirm that, within the deverbal sub-sample, the distribution of grammatical templates departs significantly from uniformity for all three prototypical cognitive sets (Table~\ref{tab:gof_deverbal}). As expected, the rejection of H\textsubscript{0} is unambiguous in every case: countable ($\chi^2(3) = 30.73$, $p < .001$), noun of action ($\chi^2(4) = 540.03$, $p < .001$), and uncountable ($\chi^2(6) = 79.65$, $p < .001$). That being said, let us see if a grammatical template is significantly favoured in comparison with other ones.

\begin{table}[H]
\centering
\caption{Chi-square goodness-of-fit results for the deverbal sub-sample, per cognitive set (H\textsubscript{0}: uniform distribution across templates).
$k$ = number of non-zero templates; $N$ = total item count.
Significance: $^{***}p < .001$.}
\label{tab:gof_deverbal}
\begin{tabular}{lrrrrl}
\hline\hline
\textbf{Cognitive set} & $N$ & $k$ & $\chi^2$ & $df$ & $p$ \\
\hline\hline
Countable      &  73 & 4 &  30.726 & 3 & $< .001^{***}$ \\
Noun of action & 159 & 5 & 540.025 & 4 & $< .001^{***}$ \\
Uncountable    &  52 & 7 &  79.654 & 6 & $< .001^{***}$ \\
\hline\hline
\end{tabular}
\end{table}

The post-hoc pairwise comparisons (Table~\ref{tab:bonferroni_deverbal}) refine this picture considerably. For the countable set ($N = 73$, $\alpha_{\text{adj}} = 0.0083$), $T_2$ emerges as the dominant template ($N = 36$, $\hat{p} = 0.493$), significantly exceeding both $T_5$ ($z = 5.27$, $p < .001$) and $T_6$ ($z = 5.05$, $p < .001$). However, the $T_1$ ($N = 22$, $\hat{p} = 0.301$) vs $T_2$ comparison does not survive correction ($p = .018 > \alpha_{\text{adj}}$), meaning that $T_1$ and $T_2$ are statistically indistinguishable as the two preferred templates for deverbal countable nouns. This constitutes a departure from the full sample pattern, where $T_1$ was unambiguously dominant, while, in the deverbal sub-sample, $T_2$ takes on a comparably prominent role, suggesting that deverbal countable nouns show a broader spread across $T_1$ and $T_2$.

\begin{table}[H]
\centering
\caption{Significant pairwise two-proportion $z$-tests with Bonferroni correction for the deverbal sub-sample. Only pairs surviving correction ($p < \alpha_{\text{adj}}$) are shown (except for $T_1$ vs $T_2$). $\hat{p}$ = observed proportion within the cognitive set; $\alpha_{\text{adj}} = 0.05/k$. Full results are provided in the Supplementary Appendix C2.}
\label{tab:bonferroni_deverbal}
\begin{tabular}{llrrrrl}
\hline\hline
\textbf{Cognitive set} & \textbf{Pair} & $\hat{p}_i$ & $\hat{p}_j$ & $z$ & $p$ & \\
\hline\hline
\multicolumn{7}{l}{\textit{Countable} ($N = 73$,\; $k = 6$,\; $\alpha_{\text{adj}} = 0.0083$)} \\
\hline
 & $T_1$ vs $T_2$ & 0.301 & 0.493 &  -2.368& $.02$   & \textit{ns}  \\
 & $T_1$ vs $T_5$ & 0.301 & 0.096 &  3.112 & $.002$   & $^{**}$  \\
 & $T_1$ vs $T_6$ & 0.301 & 0.11 &  2.868 & $.004$   & $^{**}$  \\
 & $T_2$ vs $T_5$ & 0.493 & 0.096 &  5.265 & $< .001$ & $^{***}$ \\
 & $T_2$ vs $T_6$ & 0.493 & 0.11 &  5.05 & $< .001$ & $^{***}$ \\
\hline
\multicolumn{7}{l}{\textit{Noun of action} ($N = 159$,\; $k = 10$,\; $\alpha_{\text{adj}} = 0.005$)} \\
\hline
 & $T_1$ vs $T_2$ & 0.937 & 0.019 & 16.39 & $< .001$ & $^{***}$ \\
 & $T_1$ vs $T_4$ & 0.937 & 0.006 & 16.626 & $< .001$ & $^{***}$ \\
 & $T_1$ vs $T_5$ & 0.937 & 0.019 & 16.39 & $< .001$ & $^{***}$ \\
 & $T_1$ vs $T_6$ & 0.937 & 0.019 & 16.39 & $< .001$ & $^{***}$ \\
\hline
\multicolumn{7}{l}{\textit{Uncountable} ($N = 52$,\; $k = 21$,\; $\alpha_{\text{adj}} = 0.0024$)} \\
\hline
 & $T_1$ vs $T_6$ & 0.058 & 0.558 & $-$5.524 & $< .001$ & $^{***}$ \\
 & $T_2$ vs $T_6$ & 0.173 & 0.558 & $-$4.073 & $< .001$ & $^{***}$ \\
 & $T_3$ vs $T_6$ & 0.058 & 0.558 & $-$5.524 & $< .001$ & $^{***}$ \\
 & $T_4$ vs $T_6$ & 0.115 & 0.558 & $-$4.773 & $< .001$ & $^{***}$ \\
 & $T_5$ vs $T_6$ & 0.019 & 0.558 & $-$6.06 & $< .001$ & $^{***}$ \\
 & $T_6$ vs $T_8$ & 0.558 & 0.019 &  6.06   & $< .001$ & $^{***}$ \\
\hline\hline
\multicolumn{7}{l}{\footnotesize $^{**}p < \alpha_{\text{adj}}$;\quad $^{***}p < .001$. Negative $z$: $\hat{p}_i < \hat{p}_j$.}
\end{tabular}
\end{table}


For the noun of action subgroup ($N = 159$, $\alpha_{\text{adj}} = 0.005$), the pattern is entirely consistent with the full sample. $T_1$ ($N = 149$, $\hat{p} = 0.937$) is overwhelmingly preferred and differs significantly from every other attested template (all $z \geq 16.39$, $p < .001$). All non-$T_1$ templates are statistically indistinguishable from one another, confirming that $T_1$ represents an almost categorical preference for deverbal nouns of action.

For the uncountable set ($N = 52$, $\alpha_{\text{adj}} = 0.0024$), $T_6$ ($N = 29$, $\hat{p} = 0.558$) is the sole dominant template, differing significantly from every other attested template (all $z \geq 4.07$, $p < .001$). This dominance is even more pronounced than in the full sample, where $T_6$ already stood out. In the deverbal sub-corpus, $T_6$ concentrates over half of all items, and no pair among the non-$T_6$ templates reaches significance, indicating that the remaining constructional space is largely undifferentiated.

Taken together, these findings demonstrate that the template preferences observed in the full sample are not only replicated but in some cases amplified in the deverbal sub-sample, most notably for the uncountable subgroup. This reinforces the hypothesis that cognitive set membership is a reliable predictor of grammatical template selection. For the countable set, however, a divergent pattern emerges. $T_2$ rises to a prominence comparable to $T_1$, a distribution not observed in the full sample. To confirm and extend these tendencies, the following section examines whether further partitioning of the data reveals additional meaningful subspaces.

\subsubsection{Low-dispersion subspace identification}

To corroborate the patterns identified in the previous section, we will look for other subspaces in the dataset. While the previous analysis retained only the cognitive set and deverbal variables, which proved to be robust predictors,  the identification of subspaces requires a richer feature space to operate. A reduced set of variables was therefore selected to define a lower-dimensional subspace; the variables retained are listed in Table~\ref{tab:variables}. The features of particular interest are nouns that are either exclusive to Riffian (i.e., with no known counterparts in other Berber languages) or recent loanwords. These criteria rest on the assumption that the higher the frequency and the older the word, the more likely it is to have undergone recursive morphological grammatical changes.

\begin{table}[ht!]
\centering
\caption{Description of the dataset.}
\label{tab:variables}
\begin{tabular}{@{}lp{9cm}@{}}
\hline\hline
\noalign{\smallskip}
\textbf{Variable} & \textbf{Description} \\
\noalign{\smallskip}
\midrule\midrule
\multicolumn{2}{@{}l}{\textit{Data variables}} \\
\addlinespace
Noun            & Riffian noun in orthographic notation \\
\addlinespace
Type            & Grammatical gender and number:
                  \textit{nfp.}\ (fem.\ plural),
                  \textit{nfs.}\ (fem.\ sing.),
                  \textit{nmp.}\ (masc.\ plural),
                  \textit{nms.}\ (masc.\ sing.) \\
\addlinespace
Pages           & Page number(s) in the source dictionary \citep{serhoual_dictionnaire_2002} where the noun appears \\
\addlinespace
\multicolumn{2}{@{}l}{\textit{Feature variables}} \\
\addlinespace
\multicolumn{2}{@{}l}{\quad\textit{Formula arguments}} \\
\addlinespace
Deverbal        & Whether the noun is derived from a verbal root
                  (\textit{yes}/\textit{no}) \\
\addlinespace
Riffian only    & Whether the noun is exclusive to the Riffian language (\textit{yes}/\textit{no}) \\
\addlinespace
Borrowing       & Whether the noun is a loanword from another language (\textit{yes}/\textit{no}) \\
\addlinespace
Animate         & Whether the noun refers to an animate entity (\textit{yes}/\textit{no}) \\
\addlinespace
\multicolumn{2}{@{}l}{\quad\textit{Subgroup variable}} \\
\addlinespace
Cognitive set   & Semantic/cognitive category (\textit{Countable}, \textit{Uncountable}, \textit{Noun of action}, etc.) \\
\addlinespace
\multicolumn{2}{@{}l}{\textit{Target variable}} \\
\addlinespace
Template        & Morphological template classification ($T_1$–$T_9$) \\
\noalign{\smallskip}
\hline\hline
\end{tabular}
\end{table}

We do not imply that nouns derived from old Berber verbal roots are necessarily ancient forms. It is possible, and even likely, that some deverbal nouns have been derived within the Riffian system itself (e.g., \textipa{\textit{DweR}} ‘to return’ $\looparrowright$ \textipa{\textit{Ta-DuRa}} ‘return’). However, as Riffian is mostly a spoken language, we cannot rely upon historical documentation. Hence, confirming this would require extensive analysis, including comparisons with other Berber dialects/languages, which may not yield definitive results. The presence of the same noun in different Berber languages does not necessarily prove a common origin, as identical forms could have arisen independently in different locations. Given the high risk of Type I and Type II errors, we have chosen to exclude these words.

These morphological differences are manifested at different levels. It was observed that lexemes not adhering to these criteria exhibit significant inter-word variations (see Table \ref{tab1}) and dialectal variations. Additionally, there is an abundance of deverbal nouns belonging to this problematic category, either with different morphologies for the same cognitive set and base item (e.g., Uncountable: \textipa{\textit{a-sem:i\textsubdot{D}}}, \textipa{\textit{D-esme\textsubdot{d}}} ‘cold / low temperature’ from \textipa{\textit{sme\textsubdot{d}}} ‘to be cold’) or with identical grammatical markers --- except for the gender markers --- and radical morphemes (e.g., Countable: \textipa{\textit{a-Sew:aR}} ‘harvester’ / Uncountable: \textipa{\textit{Da-Sew:aRt}} ‘harvest’ from \textipa{\textit{Sew:eR}} ‘to harvest’).

However, this generalisation does not apply to nouns of action. Remarkably, as established in Section~\ref{data_exploration}, nouns of action consistently display the same grammatical morphology, even without considering the reduced space. The vast majority bear singulative and masculine markers (e.g., \textipa{\textit{RiqReq}} ‘to twinkle’ $\looparrowright$ \textipa{\textit{a-RiqReq}} ‘fact of twinkling’; \textipa{\textit{Sfaq}} ‘stare’ $\looparrowright$ \textipa{\textit{a-Sef:aq}} ‘fact of staring’). This stability might be attributed to their infrequent use in communication. In contrast, the other nouns are used more frequently, and this higher frequency has obscured their underlying morphological pattern. Given that their template (i.e., $T_1$) was unambiguously identified in the preceding quantitative analysis, nouns of action were excluded from the training set.

By retaining only countable and uncountable nouns, the search space was reduced. Furthermore, the conditional probability $P(\text{Target} \mid \text{Subgroup})$ was redefined as the objective function. More specifically, we sought to jointly maximise the pairs of conditional probabilities $\bigl(P(T_2 \mid C)$, $P(T_6 \mid U)\bigr)$ and $\bigl(P(T_1 \mid C)$, $P(T_6 \mid U)\bigr)$.

\paragraph{The optimal filter and template: \texorpdfstring{$\max\bigl(P(T_1\!\mid\! C),\, P(T_6\!\mid\! U)\bigr)$}{max(P($T_1$|C), P($T_6$|U))} versus \texorpdfstring{$\max\bigl(P(T_2\!\mid\! C),\, P(T_6\!\mid\! U)\bigr)$}{max(P($T_2$|C), P($T_6$|U))}}
\label{sec:filter-comparison}

As established in the previous section, uncountable nouns were strongly associated with $T_6$. Consequently, we did not consider alternative templates for this cognitive set. In contrast, such a clear association was not observed for countable nouns. Therefore, our analysis pursued two objectives: (i) identifying a low-dispersion subspace for these subgroups and (ii) determining whether template $T_1$ or $T_2$ is more frequently associated with countable nouns.

\subparagraph{Preliminaries}

The present analysis evaluates four read-once formulas that operate as logical filters over the binary feature variables Deverbal ($D$), Riffian-only ($R$), Borrowing ($B$), and Animate ($A$). These formulas are constructed in two steps: feature selection, which determines which variables enter the formula, and literal selection, which assigns to each selected variable a polarity --- either a positive literal ($X = 1$) or a negative literal ($X = 0$). To meet the objectives set out above, we learn the filter that simultaneously maximises both objective functions.

\begin{enumerate}
  \item $\varphi_1$: $D{=}1 \wedge (B{=}1 \vee R{=}1) \wedge A{=}0$ \quad (OR filter)
  \item $\varphi_2$: $D{=}1 \wedge B{=}1 \wedge A{=}0$ \quad (deverbal borrowings)
  \item $\varphi_3$: $D{=}1 \wedge R{=}1 \wedge A{=}0$ \quad (deverbal Riffian-only)
  \item $\varphi_4$: $D{=}1 \wedge A{=}0$ \quad (all deverbal non-animate)
\end{enumerate}

Each filter restricts the corpus to a subgroup of Riffian nouns and induces two conditional probability estimates: $P(T_j \mid C, \varphi_i)$, the probability that a countable noun ($C$) receives morphological template $T_j$ --- that could be either $T_1~ or ~T_2$ --- under filter $\varphi_i$, and $P(T_6 \mid U, \varphi_i)$, the analogous quantity for uncountable nouns ($U$) and template $T_6$.

Two competing objective functions are compared. The first jointly optimises $P(T_2 \mid C)$ and $P(T_6 \mid U)$, targeting the template $T_2$ for countable nouns while preserving the template $T_6$. The second replaces $T_2$ by $T_1$, the morphologically unmarked masculine singular template that constitutes the default realisation for countable nouns in the unfiltered lexicon.

Joint performance across the two cognitive sets is evaluated through four complementary criteria. The \emph{harmonic mean} (H) of the two conditional probabilities penalises imbalance between the two sub-targets:
\begin{equation}
  H(\varphi_i, T_j)
  \;=\;
  \frac{2\,P(T_j \mid C, \varphi_i)\cdot P(T_6 \mid U, \varphi_i)}
       {P(T_j \mid C, \varphi_i) + P(T_6 \mid U, \varphi_i)}
  \label{eq:hmean}
\end{equation}

The \emph{lift} measures how much the filter amplifies template $T_j$ relative to the unfiltered baseline:

\begin{equation}
\mathrm{Lift}(T_j \mid S, \varphi_i) = P(T_j \mid S, \varphi_i) / P(T_j \mid S)
  \label{eq:lift}
\end{equation}

The \emph{Weighted Relative Accuracy} combines coverage and precision gain, and its sum over both cognitive sets $\mathrm{WRAcc}_{\Sigma} = \mathrm{WRAcc}(T_j, C, \varphi_i) + \mathrm{WRAcc}(T_6, U, \varphi_i)$ provides a joint coverage-weighted assessment.

\begin{equation}
  \mathrm{WRAcc}(T_j, S, \varphi_i)
  \;=\;
  \frac{n(S,\varphi_i)}{n_S}
  \cdot\bigl[P(T_j \mid S, \varphi_i) - P(T_j \mid S)\bigr]
  \label{eq:wracc}
\end{equation}

Finally, $\min\!\bigl(P(T_j \mid C),\, P(T_6 \mid U)\bigr)$ gives the worst-case guarantee over both sub-targets.

\subparagraph{Results}

Before any filter is applied, the corpus contains $n_C = 164$  and $n_U = 110$, respectively countable and uncountable nouns.  The dominant template for countable nouns is $T_1$, realised by $81$ items ($P(T_1 \mid C) = 0.494$), while $T_2$ accounts for $52$ items ($P(T_2 \mid C) = 0.317$).  For uncountable nouns, $T_6$ is the most frequent template with $40$ items ($P(T_6 \mid U) = 0.364$). These figures establish the reference against which changes introduced by the filters are assessed (see, for reproducibility, Supplementary Appendix B1–B5, which respectively process the baseline and $\varphi_1$–$\varphi_4$.).

\begin{table}[ht]
\centering
\caption{Joint optimisation of $P(T_2\mid C)$ and $P(T_6\mid U)$. Bold marks the leading value per criterion column.}
\label{tab:T2T6}
\small
\begin{tabular}{lccccccccccc}
\hline\hline
Filter & $n_C$ & $n_{T_2}$ &  $P(T_2|C)$ & $n_U$ & $n_{T_6}$ & $P(T_6|U)$ & $H$ & $\mathrm{Lift}_C$ &  $\mathrm{Lift}_U$ & $\mathrm{WRAcc}_\Sigma$ & $\min(P)$ \\
\hline\hline
$\varphi_1$ & 29 & 22 & 0.759 &  25 & 15 & 0.6 & 0.67 & 2.393 & 1.648 & 0.128 & \textbf{0.6} \\
$\varphi_2$ & 17 & 16 & \textbf{0.941} & 15  & 9 & 0.6 & \textbf{0.733} & \textbf{2.968} & 1.648 & 0.099 & \textbf{0.6} \\
$\varphi_3$ & 12 & 6 & 0.5 & 11 & 7 & \textbf{0.636} & 0.56 & 1.577 & \textbf{1.747} & 0.041 & 0.5 \\
$\varphi_4$ & 66 & 36 & 0.545 & 52 & 29 & 0.558 & 0.552 & 1.72 & 1.532 & \textbf{0.184} & 0.545 \\
Baseline    & 164 & 52 & 0.317 & 110 & 40 & 0.364 & --- & --- & --- & --- & 0.317 \\
\hline\hline
\end{tabular}
\end{table}

Table~\ref{tab:T2T6} reports all four evaluation criteria for filters $\varphi_1$--$\varphi_4$ under the first objective function. Filter $\varphi_2$ achieves the highest joint precision: $P(T_2 \mid C) = 0.941$ and $H = 0.733$, yielding a lift of $2.97$ over the unfiltered baseline for C. The $\mathrm{WRAcc}_\Sigma$ of $\varphi_2$ ($0.099$) is lower than that of $\varphi_4$ ($0.184$) because $\varphi_2$ covers fewer items; however, $\varphi_4$ achieves this larger coverage ($n_C = 66$) at the cost of substantially weaker precision ($P(T_2 \mid C) = 0.545$, $H = 0.552$). The OR filter $\varphi_1$ occupies a strategically intermediate position. It extends coverage to $n_C = 29$ at the cost of a moderate precision reduction ($P(T_2 \mid C) = 0.759$, $H = 0.67$). The worst-case guarantee $\min(P)$ is identical for $\varphi_1$ and $\varphi_2$ ($0.6$, always attained by $P(T_6 \mid U)$).

The second objective function targets $T_1$, rather than the deverbal template $T_2$. A fundamental asymmetry emerges. We can observe in Table \ref{tab:T1T6} that no filter raises $P(T_1 \mid C)$ above its baseline value of $0.494$, and the lift for $T_1$ is below $1.0$ for every filter. All filters suppress $T_1$ for C, because conditioning on $D{=}1$ selects the deverbal stratum in which $T_2$ is the dominant realisation. Additionally, the $\mathrm{WRAcc}_\Sigma$ values confirm this; three of the four filters yield a negative joint WRAcc sum, meaning the filter simultaneously reduces for countable nouns both $T_1$ precision and coverage relative to baseline. Only $\varphi_3$ achieves a marginally positive $\mathrm{WRAcc}_\Sigma = 0.009$, but this is driven entirely by the uncountable side ($\mathrm{WRAcc}(T_6, U, \varphi_3) = +0.027$ outweighing $\mathrm{WRAcc}(T_1, C, \varphi_3) = -0.018$, see Supplementary Appendix D). The best harmonic mean under objective function~2 is $H = 0.359$ ($\varphi_3$), far below the $H = 0.733$ achieved by $\varphi_2$ under objective function~1.

\begin{table}[ht]
\centering
\caption{Joint optimisation of $P(T_1\mid C)$ and $P(T_6\mid U)$. Same filter subgroups as Table~\ref{tab:T2T6}; only the Countable target template changes from $T_2$ to $T_1$. Bold marks the leading value per criterion column.}
\label{tab:T1T6}
\small
\begin{tabular}{lccccccccccc}
\hline\hline
Filter & $n_C$ & $n_{T_1}$ & $P(T_1|C)$ & $n_U$  & $n_{T_6}$ & $P(T_6|U)$ & $H$ & $\mathrm{Lift}_C$ & $\mathrm{Lift}_U$ & $\mathrm{WRAcc}_\Sigma$ & $\min(P)$ \\
\hline\hline
$\varphi_1$ & 29 & 4 & 0.138 & 25 & 15 & 0.6 & 0.224 & 0.279 & 1.65 & $-0.013$ & 0.138 \\
$\varphi_2$ & 17 & 1 & 0.059 & 15 & 9 & 0.6 & 0.107 & 0.119 & 1.65 & $-0.011$ & 0.059 \\
$\varphi_3$ & 12 & 3 & 0.25 & 11 & 7 & \textbf{0.636} & \textbf{0.359} & \textbf{0.506} & \textbf{1.75} & \textbf{0.009} & 0.25 \\
$\varphi_4$ & 66 & 16 & 0.242 & 52 & 29 & 0.558 & 0.338 & 0.491 & 1.534 & $-0.01$ & 0.242 \\
Baseline    & 164 & 81 & \textbf{0.494} & 110 & 40 & 0.364 & --- & --- & --- & --- & \textbf{0.364} \\
\hline\hline
\end{tabular}
\end{table}

\paragraph{Discussion}

The two objective functions define qualitatively different landscapes. Under objective function~1, precision gain for $T_2 \mid C$ and for $T_6 \mid U$ are complementary. The morpho-lexical features that concentrate $T_2$ among countable nouns simultaneously select an uncountable subpopulation clustering in $T_6$, so tightening the filter raises both sub-target probabilities. Every evaluation criterion ranks $\varphi_2$ first when precision is prioritised, and $\varphi_4$ first when joint coverage is the primary concern.
 
Under objective function~2, the two sub-targets are in structural conflict; deverbal nouns systematically deviate from $T_1$ by
favouring $T_2$. Consequently, optimising $P(T_1 \mid C)$ within the present filter space is not a coherent empirical objective. 

In conclusion, the best solution for countable nouns is filter $\varphi_2$ under objective function~1:
\[
  P(T_2 \mid C,\,\varphi_2) = 0.941,\quad
  P(T_6 \mid U,\,\varphi_2) = 0.6,\quad
  H(\varphi_2,\,T_2) = 0.733.
\]

This result admits a coherent linguistic interpretation. The complementarity of $T_2$ (Countable) and $T_6$ (Uncountable) within $\varphi_2$ reflects a systematic morphological alternation governed by the countability paradigm $[\pm\mathrm{SING}, \pm\mathrm{COL}]$, making $\varphi_2$ the most informative joint discriminator for both cognitive sets. This finding also supports the hypothesis that the patterns observed in the full dataset are the outcome of diachronic lexical change. Most of the nouns concerned have undergone successive derivatives, producing the morphological divergences that underlie the observed distribution. 

It is worth noting that, in the full sample as well as in the subspace of deverbal nouns (filter $D = 1$), $T_1$ showed a consistent and significant association with countable nouns. Recall that $T_1$ and $T_2$ differ only in their gender features. The predominance of $T_1$ therefore suggests that grammatical gender shifts in Riffian are productive. Moreover, a substantial proportion of the nouns assigned to $T_1$ appear to be derived from deverbal base nouns associated with $T_2$, indicating a systematic gender shift rather than a random assignment of templates. This interpretation is further supported by the non-uniform distribution of $C$ discussed in Section \ref{data_exploration}. This finding further supports the existence of a constraining mechanism governing both the direction of lexical change and template selection, as will be established in Section \ref{model_CN}.

The OR filter $\varphi_1$ remains a viable alternative when coverage is prioritised. It increases the number of correctly classified countable items from $16$ to $22$ while keeping both $P(T_6 \mid U)=0.6$ and the worst-case guarantee $\min(P)=0.6$ unchanged. The loss in precision (i.e., a larger dispersion) is moderate ($H$ decreases from $0.73$ to $0.67$), indicating that the precision--coverage trade-off between $\varphi_1$ and $\varphi_2$ remains within a solution space that is already jointly optimal with respect to the $T_2$--$T_6$ opposition. No filter targeting $T_1$ for countable nouns can achieve a comparable balance.

To draw conclusions about the relationship between cognitive sets and grammatical templates, an omnibus test of independence was performed on the subspace obtained from $\varphi_1$. Because the subspace associated with $\varphi_2$ is nested within that of $\varphi_1$, any association detected in the broader $\varphi_1$ space necessarily reflects patterns present in the data available to $\varphi_2$. We therefore treat the $\varphi_1$ analysis as a baseline assessment of the cognitive set-template relationship, while recognising that the strength of the association within the more restricted $\varphi_2$ subspace may differ. Templates $T_3$ and $T_7$ contained zero observations in both cognitive sets and were therefore excluded from the analysis to avoid distorting the degrees of freedom. Because 67\% of cells had expected frequencies below 5, significance was assessed using a Monte Carlo version of Pearson's chi-square test with 100{,}000 replicates rather than the asymptotic $\chi^2$ distribution. The observed statistic was $\chi^2=31.06$, and no simulated table produced a value as extreme as the observed statistic ($p < 10^{-5}$). A Monte Carlo Fisher's exact test yielded the same result ($p < 10^{-5}$), confirming a highly significant association between cognitive set and template selection, indicating that the two subgroups exhibit fundamentally distinct structural profiles.

This finding likewise supports treating the semantic features C, U, and NA as parameters of the gradient function $g$. Therefore, this method can also be used to learn the elements of the tuple $k$ $\in$ $K$ (see Supplementary Appendix A). More broadly, any feature variable that defines a subgroup variable and that exhibits a sufficiently strong association with a target variable may serve as a candidate parameter. This reduction technique may provide a systematic, learning-based approach to discovering model parameters.

\subsubsection{Subspace and candidate extraction}

We use the filter $\varphi_2$, which has the lowest dispersion, to reduce the full space to a subspace. After excluding these words to minimise obscure morphological patterns, we obtain a smaller and cohesive subspace to ascertain the initial templates of each cognitive set. As expected from this dimension reduction method, the subspace size is not large. This outcome stems from the nature of these words. Recent verbal lexemes tend to produce fewer derived forms, and when they do, they are typically nouns of action (e.g., \textipa{\textit{Reb:eT}} ‘be silent’ $\looparrowright$ \textipa{\textit{a-Reb:eT}} ‘fact of being silent’). Moreover, since U and C generally correspond to concrete and abstract meanings, these elements are often directly borrowed from other languages, primarily Romance and Arabic.

\begin{table}[h!]
  \centering
  \caption{Subset of the subspace for obtaining initial templates of C and U}
  \resizebox{15.5cm}{!}{%
   \small
    \begin{tblr}{
      width = \linewidth,
      colspec = {Q[350]Q[20]Q[300]Q[11]Q[300]Q[20]Q[350]},
      hline{1} = {-}{black},
      hline{1} = {2}{-}{black},
      hline{2} = {-}{black},
      hline{2} = {2}{-}{black},
      hline{9} = {-}{black},
      hline{9} = {2}{-}{black},
      }
      \SetCell[c=3]{c} Deverbal countable noun   & 1-2~    &       1-3~            &        &  \SetCell[c=3]{c} Deverbal uncountable noun  &       1-6         &  1-7  \\

\textipa{\textit{BuR}} ‘to micturate’ & $\looparrowright$ & \textipa{\textit{Da-Bew:aRt}} ‘bladder’ & ~ & \textipa{\textit{dhen}} ‘to grease’ & $\looparrowright$ & \textipa{\textit{dduhneT}} ‘unctuousness’ \\
      
\textipa{\textit{feSfeS}} ‘to be effervescent’ & $\looparrowright$ & \textipa{\textit{Da-feSfaSt}} ‘spray’ & ~ & \textipa{\textit{lul:eS}} ‘to sparkle’ & $\rightarrow$ & \textipa{\textit{t-lul:eSt}} ‘toy’ \\

\textipa{\textit{\textinvscr ij:es}} ‘to be muddy’ & $\looparrowright$ & \textipa{\textit{Da-\textinvscr ij:ast}} ‘mire’ & ~ & \textipa{\textit{bumbes}} ‘to get dark’ & $\looparrowright$ & \textipa{\textit{t-bambast}} ‘twilight’ \\

\textipa{\textit{new:eR}} ‘to bloom’ & $\looparrowright$ & \textipa{\textit{Da-new:aRt}} ‘flower’ & ~ & \textipa{\textit{d:a\textcrh a}} ‘to be proud’ & $\looparrowright$ & \textipa{\textit{d:a\textcrh iT}} ‘pride’ \\

\textipa{\textit{sef:aR}} ‘to whistle’ & $\rightarrow$ & \textipa{\textit{Da-sef:aRt}} ‘whistle’ & ~ & \textipa{\textit{fTeR}} ‘to roll semolina’ & $\looparrowright$ & \textipa{\textit{T-faTaRt}} ‘semolina’ \\
       
\textipa{\textit{sij:eq}} ‘to mop the floor’ & $\looparrowright$ & \textipa{\textit{Da-sij:aqt}} ‘squeegee’ & ~ & \textipa{\textit{dek:}} ‘to sip’ & $\rightarrow$ & \textipa{\textit{dek:eT}} ‘sip’ \\
       
\textipa{\textit{\textcrh eRB}} ‘to fan’ & $\looparrowright$ & \textipa{\textit{Da-\textcrh eR:aBt}} ‘fan’ & ~ & \textipa{\textit{\textrevglotstop uf:en}} ‘to be dirty’ & $\rightarrow$ & \textipa{\textit{D-u\textrevglotstop uf:na}} ‘horridness’ \\

    \end{tblr}\label{sample}
  }
\end{table}

Nevertheless, the filtered sample reveals candidate templates with significant frequency emerging for each cognitive set, as seen earlier. In Table \ref{sample}, a subset of the subspace containing those frequent candidate templates is listed for U and C. While other templates occasionally appear in the subspace, they do so infrequently, only once or twice for filter $\varphi_2$. These false positives can be attributed to various factors, including transcription errors, informant errors, idiolectal variation, or even reverse derivational patterns (i.e., noun to verb, see Figure \ref{fig_new}.π).

To select the initial templates from this subspace, we computed the statistical mode for each subgroup. This procedure is analogous to the majority-voting mechanism employed in several machine-learning classification frameworks \citep{Hastie2009}. For example, the $k$-Nearest Neighbour (k-NN) classifier typically assigns a class label according to the modal class among the nearest neighbours, while many ensemble classifiers, such as Random Forests, determine their final prediction through majority voting across individual learners. This analogy clarifies the selection logic. A separate question, however, concerns the robustness of the selected template. With a view to generalising the proposed methodology to other lexical domains or languages, it is important to ensure that, in general, the selected target remains sufficiently representative even when its relative frequency is lower than that observed in the present study. A criterion is therefore called for assessing the robustness.

To this end, we draw on the intuition underlying the Condorcet Jury Theorem --- a result standardly invoked alongside majority-voting methods \citep{Boland1989Condorcet} ---, according to which a majority decision is reliable when each voter has a probability of success greater than 0.5. By analogy, we regard a target as a valid candidate when its relative frequency exceeds 0.5 within the optimised subgroup. This condition is satisfied by both $T_2$ and $T_6$ (see Table \ref{tab:T2T6}), thereby supporting the validity of the selected templates. Applying this method, we are able to establish the initial templates (see Table~\ref{template_init}). The initial template for countable nouns is therefore characterised by feminine and singulative markers. In contrast, the initial template for uncountable nouns also carries the feminine marker but exhibits a collective marker.

\begin{table}[h!]
\aboverulesep = 0.1cm
\belowrulesep = 0.1cm
\centering
\caption{Initial templates of each cognitive set}
\begin{tabular}{ll} 

\toprule\toprule

\multicolumn{1}{c}{Cognitive set} & \multicolumn{1}{c}{Initial template}\\ 

\midrule\midrule

Countable                 & $T_2$: $\{N,+SG,-PL,-M,+F,-COL,+SING\}$  \\ \rule{0pt}{0.4cm}\noindent
Uncountable             & $T_6$: $\{N,+SG,-PL,-M,+F,+COL,-SING\}$  \\ \rule{0pt}{0.4cm}\noindent
Noun of action          & $T_1$: $\{N,+SG,-PL,+M,-F,-COL,+SING\}$  \\ 

\bottomrule\bottomrule

\end{tabular}
\label{template_init}
\end{table}

Having established our theoretical and methodological framework, we will now present the detailed findings derived from our analysis.

\subsection{Formal modelling of gender shift dynamics}\label{model_CN}

This section aims to explore the factors that explain why some nouns undergo gender shifts while others do not. To address this, we must delve into the processes of word and meaning formation that these nouns undergo. The parts of speech analysed include verbs and nouns, with transitions between categories occurring through word formation (morphological derivation or conversion) or semantic change. Figure \ref{fig_new} uses symbol functions to trace the derivational direction (left-to-right or right-to-left) from one category to another. We identified eleven distinct derivational chains followed by various nouns. For each of them, examples are provided as illustrations. However, these examples are just instances of a broader and more complex word and meaning formation model. This study primarily emphasises the directional derivatives, outlined in Figure \ref{fig_new}, that lead to grammatical gender shifts.

\begin{figure}[h!]
  \renewcommand{\thefigure}{\arabic{figure}}

  \centering
  \caption{Formation model of gender in Riffian}
  \vskip 0.2cm

  \resizebox{\linewidth}{!}{%
    \small
    \begin{tabular}{l l l l l l}
$\alpha$. & V                             & $\looparrowright$ & NA                                    & $\supseteq$ & U                                             \\
      ~  & \textipa{\textit{\textsubdot{R}aZa}} ‘to wait’      & ~   & \textipa{\textit{a-\textsubdot{R}aZi}} ‘act of waiting’     & ~   & \textipa{\textit{a-\textsubdot{R}aZi}} ‘wait’                                        \\
   ~  & ~ 	                                & ~  &  (\textipa{\textit{\textsubdot{R}aZi}},  $\{N,+SG,-PL,+M,-F,-COL,+SING\}$)   & ~   & (\textipa{\textit{\textsubdot{R}aZi}},  $\{N,+SG,-PL,+M,-F,-COL,+SING\}$)      \\
         ~  & ~ 	                                & ~  & ~                                     & ~   & ~                                                             \\

         $\mu$. & V                          & $\looparrowright$  & NA                                     & $\rightarrow$ & U                                             \\
      ~  & \textipa{\textit{sendu}} ‘to churn’   & ~   & \textipa{\textit{a-sendu}} ‘act of churning’               & ~   & \textipa{\textit{Da-senduT}} ‘cream’                                    \\
          ~  & ~ 	                                & ~  & (\textipa{\textit{sendu}},  $\{N,+SG,-PL,+M,-F,-COL,+SING\}$)   & ~   & (\textipa{\textit{sendu}},  $\{N,+SG,-PL,-M,+F,-COL,+SING\}$)      \\
         ~  & ~                             & ~   & ~                                     & ~   & ~                                                             \\

         $\gamma$. & V                             & $\rightarrow$ & NA                                    & ~   & ~                                                             \\
      ~  & \textipa{\textit{sum:eR}} ‘to sunbathe’ & ~   & \textipa{\textit{a-sum:eR}} ‘act of sunbathing’ & ~   & ~                                                             \\
      ~  & ~ 	                                & ~  & (\textipa{\textit{sum:eR}},  $\{N,+SG,-PL,+M,-F,-COL,+SING\}$)   & ~   & ~\vspace{0.1cm} \\
      ~  & ~                             & $\looparrowright$ & U                                     & $\rightarrow$ & C                                                             \\
      ~  & ~                             & ~   & \textipa{\textit{t-sam:eRT}} ‘sunny place’      & ~   & \textipa{\textit{sam:eR}} ‘south-facing slope’                           \\
      ~  & ~                             & ~  & (\textipa{\textit{sam:eR}},  $\{N,+SG,-PL,-M,+F,+COL,-SING\}$)   & ~   & (\textipa{\textit{sam:eR}},  $\{N,+SG,-PL,+M,-F,+COL,-SING\}$)      \\
      ~  & ~                             & ~   & ~                                     & ~   & ~                                                             \\

         $\delta$. & V                             & $\rightarrow$ & U                                     & $\supseteq$ & NA                                                            \\
      ~  & \textipa{\textit{i\textinvscr is}} ‘to be smart’  & ~   & \textipa{\textit{D-i\textinvscr ist}} ‘intelligence’       & ~   & \textipa{\textit{D-i\textinvscr ist}} ‘fact of being intelligent’                   \\
      ~  & ~                             & ~  & (\textipa{\textit{i\textinvscr is}},  $\{N,+SG,-PL,-M,+F,+COL,-SING\}$)   & ~   & (\textipa{\textit{i\textinvscr is}},  $\{N,+SG,-PL,-M,+F,+COL,-SING\}$)     \vspace{0.1cm} \\
      ~  & ~                             & $\looparrowright$ & U                                     & $\rightarrow$ & C                                                             \\
      ~  & ~                             & ~   & \textipa{\textit{mi\textinvscr is}} ‘clever one’           & ~   & \textipa{\textit{mi\textinvscr is}} ‘clever’                                       \\
      ~  & ~                             & ~  & (\textipa{\textit{mi\textinvscr is}},  $\{N,+SG,-PL,+M,-F,+COL,-SING\}$)   & ~   &  (\textipa{\textit{mi\textinvscr is}},  $\{N,+SG,-PL,+M,-F,+COL,-SING\}$)      \\
      ~  & ~                             & ~   & ~                                     & ~   & ~                                                             \\

         $\epsilon$. & V                             & $\looparrowright$ & U                                     & $\rightarrow$ & C                                                             \\
      ~  & \textipa{\textit{uDRus}} ‘to be rare’  & ~   & \textipa{\textit{muDRus}} ‘precious one’        & ~   & \textipa{\textit{muDRus}} ‘slim’                                       \\
      ~  & ~                             & ~  & (\textipa{\textit{muDRus}},  $\{N,+SG,-PL,+M,-F,+COL,-SING\}$)   & ~   &  (\textipa{\textit{muDRus}},  $\{N,+SG,-PL,+M,-F,+COL,-SING\}$)      \\
      ~  & ~                             & ~   & ~                                     & ~   & ~                                                             \\

         $\zeta$. & ~                             & ~   & C                                     & $\rightarrow$ & U                                                             \\
      ~  & ~                             & ~   & \textipa{\textit{a-Rgaz}} ‘man’                 & ~   & \textipa{\textit{Da-Rgazt}} ‘courage’                                    \\
      ~  & ~                             & ~  & (\textipa{\textit{Rgaz}},  $\{N,+SG,-PL,+M,-F,-COL,+SING\}$)   & ~   &  (\textipa{\textit{Rgaz}},  $\{N,+SG,-PL,-M,+F,-COL,+SING\}$)      \\  
      ~  & ~                             & ~   & ~                                     & ~   & ~                                                             \\

         $\eta$. & V                             & $\looparrowright$ & NA                                    & ~   & ~                                                             \\
      ~  & \textipa{\textit{ndeh}} ‘to drive’      & ~   & \textipa{\textit{a-ndah}} ‘act of driving’      & ~   & ~                                                             \\
      ~  & ~                             & ~  & (\textipa{\textit{ndah}},  $\{N,+SG,-PL,+M,-F,-COL,+SING\}$)   & ~   &  ~    			\vspace{0.1cm}	\\  
      ~  & ~                             & $\looparrowright$ & C                                     & $\supseteq$ & U                                                             \\
      ~  & ~                             & ~   & \textipa{\textit{Da-nedhiwT}} ‘way of driving’            & ~   & \textipa{\textit{Da-nedhiwT}} ‘driving’                                 \\
      ~  & ~                             & ~  & (\textipa{\textit{nedhiw}},  $\{N,+SG,-PL,-M,+F,-COL,+SING\}$)   & ~   &  (\textipa{\textit{nedhiw}},  $\{N,+SG,-PL,-M,+F,-COL,+SING\}$)      \\  
      ~  & ~                             & ~   & ~                                     & ~   & ~                                                             \\

         $\lambda$. & ~                             & $\Rsh$ & U                                     & $\supseteq$ & C                                                             \\
      ~  & ~                             & ~   & \textipa{\textit{t-s:a\textrevglotstop eT}} ‘hour’               & ~   & \textipa{\textit{t-s:a\textrevglotstop eT}} ‘watch’                                      \\
      ~  & ~                             & ~  & (\textipa{\textit{s:a\textrevglotstop}},  $\{N,+SG,-PL,-M,+F,+COL,-SING\}$)   & ~   &  (\textipa{\textit{s:a\textrevglotstop}},  $\{N,+SG,-PL,-M,+F,+COL,-SING\}$)      \\  
      ~  & ~                             & ~   & ~                                     & ~   & ~                                                             \\

         $\rho$. & ~                             & ~   & C                                     & $\looparrowright$ & U                                                             \\
      ~  & ~                             & ~   & \textipa{\textit{a-k:u\textcrh}} ‘small’               & ~   & \textipa{\textit{k:uk:u\textcrh}} ‘tiny one’                                    \\
      ~  & ~                             & ~  & (\textipa{\textit{k:u\textcrh}},  $\{N,+SG,-PL,+M,-F,-COL,+SING\}$)   & ~   &  (\textipa{\textit{k:uk:u\textcrh}},  $\{N,+SG,-PL,+M,-F,+COL,-SING\}$)      \\
      ~  & ~                             & ~   & ~                                     & ~   & ~                                                             \\

         $\nu$. & ~                             & $\Rsh$ & U                                     & $\rightarrow$ & C                                                             \\
      ~  & ~                             & ~   & \textipa{\textit{ReS:in}} ‘oranges’             & ~   & \textipa{\textit{Da-ReS:int}} ‘orange’                                  \\
      ~  & ~                             & ~  & (\textipa{\textit{ReS:in}},  $\{N,+SG,-PL,+M,-F,+COL,-SING\}$)   & ~   &  (\textipa{\textit{ReS:in}},  $\{N,+SG,-PL,-M,+F,-COL,+SING\}$)      \\
      ~  & ~                             & ~   & ~                                     & ~   & ~                                                             \\

         $\pi$. & V                             & $\leftarrow$ & U                                     & ~   & ~                                                             \\
      ~  & \textipa{\textit{faD}} ‘to be thirsty’    & ~   & \textipa{\textit{faD}} ‘thirst’                & ~   & ~                                                             \\
      ~  & ~                                     & ~  & (\textipa{\textit{faD}},  $\{N,+SG,-PL,+M,-F,+COL,-SING\}$)   & ~   &  ~  \\
   
    \end{tabular}
  }
  \label{fig_new}
\end{figure}

Therefore, while additional patterns could have been included, they fall beyond the scope of this study. We set aside the broader modelling of word and meaning formation in Riffian to concentrate specifically on the mechanisms underlying gender shift. Despite some processes originating from a common input head, the ensuing iterative transfer operations (i.e., $h(w)$) redirect items toward different syntactic categories and morphological templates. To emphasise these distinctions, we have defined disjoint functions for each process.

Processes such as ε, γ, μ, η, α, and δ specifically require a verb to serve as the starting category. While α and μ share similar first derivatives resulting in nouns of action, their second derivatives diverge regarding their template (i.e., a gender shift occurs with μ but not with α). Processes δ and ε involve different word formations. δ uses conversion from V to U, while ε creates a noun of address by employing morphological derivation (i.e., the derivational marker \textit{m-}). Even though the same word formation is used, namely morphological derivation, the processes γ and ε produce different types of U, necessitating their separation. The output of process ε is a noun of address, distinct from the uncountable noun generated by process γ. The process η is meant to illustrate the case of deverbal C. Unlike γ or ε, process π involves a transition from U to V, not vice versa. This conclusion is supported by the absence of any countability markers.

Unlike other processes, the processes λ, ρ, and ζ do not synchronically derive from any known verb. That said, each exhibits distinct characteristics. For example, process ρ is a morphological derivation that derives a noun of address from an adjective. Although processes ν and λ share the same initial input conditions ---namely borrowing and uncountable--- they are treated as separate processes. Their first derivatives will produce different types of nouns and, as they are loanwords, their gender will be calqued on that of the donor language. In comparison, such characteristics are absent from the other processes. ρ and ζ share similarities in their approach to lexical derivation (i.e., a transition from C to U); however, the U created in ρ has the masculine marker since it refers to a noun of address.

To gain a deeper understanding of the morphological system, we will examine deverbal NAs (e.g., processes α and μ). These nouns typically have masculine markers, while feminine markers are associated with C and U. This raises the question of why masculine markers appear with NA, even though NA and U share many semantic similarities (e.g., non-discretisable). Despite this, the masculine marker is still prefixed.

This null morpheme serves as the default marker, that is to say, it is the unmarked form of this cognitive set, as are the singular and singulative meanings. As a member of the N syntactic category, its unmarked template includes paradigmatic markers that cannot be ignored. The choice of marker is not solely determined by semantic properties, as semantics-driven approaches suggest, but is also influenced by the item's affiliation to a cognitive set.

In summary, according to our theory, every item is assigned to a template with predefined grammatical features. Due to this default assignment, an item will carry a grammatical feature even if its semantic properties do not align with that grammatical meaning. Thus, it is the absence of this characteristic, along with dynamic template shifts (see Section \ref{shift}), that defines semantics-driven theories. Since semantics-driven theories are defined negatively, they encompass a broad range of theoretical frameworks, spanning Generativism \citep{DistributedMorphologyToday2013}, traditional descriptive linguistics \citep{greenberg2010language}, Psycholinguistics \citep{spivey2012cambridge}, and Cognitive Linguistics \citep{onysko2010cognitive}.

Figure \ref{fig_new} also highlights features distributed throughout the model, particularly the gender shifts that occur following lexical derivation (e.g., process ζ, see also (\ref{problem}) and (\ref{problem2}) in \nameref{intro}). These features will be discussed in detail in the next section.

\subsection{Template shift: Gender shift and non-gender shift}\label{shift}

We would like to begin this section by introducing the traditional views on gender shift in the literature. This linguistic phenomenon should be examined from two perspectives: why some derived nouns shift in gender, while others remain unaffected. When state variation occurs (i.e., lexical change), it is essential to propose a model that can provide a rationale for both the non-stationary state (e.g., gender shift) and the stationary state (e.g., non-gender shift). While the first question is often addressed, the second is rarely, if ever, considered.

As illustrated before, gender shift coincides with derivation. Rival morphological theories are presently unable to fully account for this observation. This raises the question of the scope of lexical derivation. In the view of some researchers \citep{fassi_fehri_constructing_2018}, the feminine marker functions as a derivational morpheme. According to this view, conversion should produce an output that matches the input form exactly. This approach represents another instance of the faithfulness presumption, but here it involves a form-to-form correspondence rather than a meaning-to-form correspondence. This assumption has led some researchers to classify such phenomena as morphological derivation, which typically does not preserve the original morphology.

The latter position should also be considered alongside linguistic descriptions that have noted the presence of gender-based derivational markers, particularly feminine ones (e.g., in German with \textit{-ung: zahl-en} ‘to pay’ / \textit{die Zahl-ung} ‘payment.F’, with \textit{-schaft: Bürger} ‘citizen.M’ / \textit{Bürger-schaft} ‘citizenry.F’ \citep[p.105]{lieber2016}). Hence, from a cross-linguistic perspective, it is unlikely that two such derivational markers would be used simultaneously\footnote{That is to say, within the same word, the use of a feminine marker as a derivational marker and the use of a gender-based derivational marker; as seen for example in Italian: \textit{fratell-anz-a} ‘fraternity.F’ from \textit{fratello} ‘brother.M’, where \textit{anz} is a gender-based derivational marker, \textit{a} is a feminine marker \citep[p. 65]{dardano1978}.}. Furthermore, it will be superfluous to assume that the grammatical template can be assigned multiple times, once to the root and once to the affix. These observations, combined with the possibility of gender shift without overt derivational markers (e.g., in French: \textit{le voile} ‘veil.M’/ \textit{la voile} ‘sail.F’, \textit{le mort} ‘dead person.M’/ \textit{la mort} ‘death.F’ \citep{Ullmann1953}), provide strong evidence for the separation of derivation and inflection.

Another implication of the TBMC model is that morphological derivation is strictly a process of lexical creation (see Section \ref{method}). Consequently, we reject theories that posit derivational relationships between animate nouns with different gender markers but the same radical --- for example, the French pair \textit{le lion} ‘the lion’ and \textit{la lionne} ‘the lioness’ --- and, more importantly, the claim that gender itself participates in this derivation \citep{Doleschal2015}. Our primary objection\footnote{We do not deny that some such pairs can result from derivation (e.g., in French: \textit{mule} ‘female mule’ and \textit{mulet} ‘male mule’; \textit{mulet} derives from \textit{mul} with the derivational suffix \textit{-et} \citep[p.491]{Dauzat1980}). However, in this case, we are dealing with a case of suppletion.} concerns the computational assumptions underlying such theories ---namely, that gender is handled by a dedicated component separate from those managing other grammatical features. In contrast, we argue that grammatical gender cannot be treated as peripheral to the template: it is necessarily integrated within it. Accordingly, we also assume that inflection, rather than derivation, underlies such morphological variations (cf. \cite{spencer2002}).

In our theoretical perspective, templates and items are distinctly separated, with no overlap between them. Therefore, we disagree with approaches that suggest otherwise. We refer to various studies that utilise the concept of ‘derivational gender affixes/markers/morphemes’ or ‘feminine derivational affixes/markers/morphemes’ (e.g., \cite{aikhenvald2016}). Instead, it is argued that the morphological system assigns a new template to the resultant items, which may carry different or identical grammatical meanings. This mechanism is what leads to gender shift. The following sections will provide further details on this process.

\subsubsection{Lexical changes: Discussion and theoretical considerations}

Lexical changes between the resultant and base items can occur through morphological derivation or semantic change. For the sake of clarity, let us first focus solely on the case of conversion. Conversion results in the formation of a distinct lexical entry; that is, each instance of conversion introduces an additional item into the set $W$ (see Proposition~\ref{pro2}). Furthermore, conversion affects the semantic properties of the item and the template. However, a template shift is expected to occur only under specific conditions. The feminine marker is typically prefixed to derived nouns only when the base noun is masculine. These base nouns can be C (see Figure \ref{fig_new}.ζ), U (see Figure \ref{fig_new}.ν) or NA (see Figure \ref{fig_new}.μ). Another interesting observation is that feminine gender marking can either appear or disappear during conversion.

Thus, if the starting category carries the feminine marker, it implies its absence in the derived category (see Figure \ref{fig_new}.γ). We refer to this as \textsc{inverse gender marking}. However, for nouns with animate referents, conversion also occurs without gender shift (see Figure \ref{fig_new}.δ). Let us recall that these markers remain inflectional and the process leading to the template shift operates outside the bounds of traditional morphological derivation. To further substantiate this, it is observed that gender shifts may accompany opposing category changes, from a countable noun to an uncountable one (C $\rightarrow$ U) and vice versa. For example, the lexeme \textipa{\textit{a-gez:aR}} ‘butcher’, a countable noun, can become \textipa{\textit{Da-gez:aRT}} ‘butchery’, an uncountable one. Likewise, the uncountable noun \textipa{\textit{RXuX}} ‘plums’ can become a countable one, \textipa{\textit{Da-RXuXt}} ‘a plum’. Accordingly, the notion that a single derivational marker could perform two contradictory operations is untenable. This contradiction motivated our proposal of a better-suited word formation process.

Our previous analysis focused on conversion, but similar observations can be made about morphological derivation. Derivational markers, whether used in inter-lexical or intra-lexical derivation, such as the French \textit{-té} or German \textit{-schaft} and \textit{-heit}, can also lead to gender shift or, more broadly, template shifts that include grammatical gender meaning. While some propose that grammatical feminine meaning is associated with these markers, this hypothesis lacks evidence, as explained earlier (see Section \ref{shift}). We propose, rather, that these instances also involve a template shift. We suggest that derivational markers are grammatically neutral, which we may name \textsc{grammaticalness-blind markers}, while inflectional markers carry grammatical meanings.

In other words, we advocate for the \textsc{integrity principle} of the grammatical template. A well-formulated item-template pair must be of the form $(w, t)$,  for instance: (\textit{zahlung},  $\{N,+SG,-PL,-M,+F,-COL,+DEF\}$). This means that no word form is assigned multiple templates with identical or asymmetrical meanings. By contrast, item-template pairs violating the principle of integrity would include irregular pairings such as: [(\textit{Zahl},  $\{N,+SG,$ $-PL,-COL,+DEF\}$), (\textit{ung},  $\{-M,+F\}$)] or [(\textit{Zahl},  $\{N,+SG,-PL,-M,+F,-COL,$ $+DEF\}$), (\textit{ung},  $\{N,+SG,-PL,-M,+F,-COL,+DEF\}$)]. Thus, we argue that verb-to-noun derivational markers do not bear grammatical meanings in the same way as those used in noun-to-noun derivation. It would be inconsistent to treat this feature as a particularity of intra-lexical derivational markers alone; rather, we must provide a unified analysis. This decision to adopt a unified account is guided by the principle of parsimony. Although the template shift of inter-lexical derivation is not conditioned by the gradient condition of the gender shift (see \emph{infra}), a template shift nonetheless occurs and is based on an initial template (see Section \ref{init_temp}).

\subsubsection{Learning set parameters: Gradient conditions}

In summary, when new nouns are derived from existing nouns, regardless of their affiliation with a cognitive set, their gender may shift, but only under specific conditions. If either the base noun or the derived noun refers to an animate object, a gender shift is consistently restricted. Otherwise, the gender shift occurs. As outlined in our theory, to capture this dynamic grammatical shift, we rely on the gradient condition algorithm (see Definition \ref{def2}), which is trained on observed data. During training, the gradient condition returns a set parameter $p$.

Because the phenomenon under investigation exhibits both stationary and non-stationary states, two set parameters are required to model the observed pattern. Accordingly, grammatical gender shift is determined by gradient condition \ref{gradcond2}. From a linguistic perspective, this mechanism contributes to the variability of morphological patterns within each cognitive set, that is, to inter-word variation (see Table \ref{tab1}).

\begin{equation}
  \begin{split}
   \resizebox{0.2\textwidth}{!}{Gradient condition \ref{gradcond2}:} & ~ \resizebox{0.66\textwidth}{!}{$\exists w \in W$, if $\exists t_i$ such that $t_i \subset k = f(w) \in K, N \in t_i$, and if $\exists t_j$ such that} \\
 & ~ \resizebox{0.61\textwidth}{!}{$t_j = h(w) \in T, N \in t_j$, then their symmetric difference must be}                 \\
 & ~ \resizebox{0.28\textwidth}{!}{$t_i~ \Delta ~t_j = \{-F, +M, +F, -M\}$}                                                                            \\
  \end{split}
  \label{gradcond2}
\end{equation}

Gradient condition \ref{gradcond2} addresses the non-stationary state case; we now turn to discuss the stationary state. This pattern is observed when semantic change is involved, especially semantic widening (see Proposition~\ref{pro2}). This difference in gender shift becomes evident through the partial functions α and μ (see Figure \ref{fig_new}). While the resultant items from the second derivative belong to the same cognitive set U, their gender differs. This distinction is a result of word and meaning formation operating between category changes. In process α, word formation is employed, hence gradient condition \ref{gradcond2} is in charge of assigning the template. As to semantic widening used in process μ, another gradient condition is responsible for pairing the resultant item with a template. This is carried out through the gradient condition \ref{gradcond3}.

\begin{equation}
  \begin{split}
\resizebox{0.2\textwidth}{!}{Gradient condition \ref{gradcond3}:} & ~ \resizebox{0.66\textwidth}{!}{$\exists w \in W$, if $\exists t_i$ such that $t_i \subset k = f(w) \in K, N \in t_i$, and if $\exists t_j$ such that} \\
 & ~ \resizebox{0.61\textwidth}{!}{$t_j = h(w) \in T, N \in t_j$, then their symmetric difference must be}                     \\
 & ~ \resizebox{0.1\textwidth}{!}{$t_i~ \Delta ~t_j = \{\}$}                                                                              \\
  \end{split}
  \label{gradcond3}
\end{equation}

When semantic widening is selected as the lexical change, the gradient condition \ref{gradcond3} must return an empty set. Consequently, no template shift occurs (neither gender shift nor any grammatical shift), and the resultant item inherits the same template as the base item.

\subsubsection{Predictive inference: Formal considerations and generalisation}

Therefore, gender shift and non-gender shift are deterministically governed, and both are outcomes of $h(w)$. To demonstrate this point, let us consider an unseen example in French and use the TBMC model for prediction. Let us assume that a partial function $h_\xi$ maps the item \textit{gland}\textsubscript{1} to the template $\{\text{N}, +\text{SG}, -\text{PL}, +\text{M}, -\text{F}, +\text{DEF}, -\text{COL}\}$. The first directional derivative of the item \textit{gland}\textsubscript{1} along the countable meaning produces a new item: ${\rightarrow}(\text{\textit{gland}}_1) = \text{\textit{gland}}_2$. The derived item is subsequently associated with a template via $h_\xi$. In this step, the item \textit{gland}\textsubscript{2} is first paired with an element of $K$:

\begin{equation}
f_\xi(\text{\textit{gland}}_2) = \{ \text{CONV}, \{ \text{N}, +\text{SG}, -\text{PL}, +\text{M}, -\text{F}, +\text{DEF}, -\text{COL} \}, \text{C}, \text{\textit{gland}}_1 \}
\end{equation}

Then, the latter output is paired with the template, resulting in the word-form \textit{la glande} with a different gender from the base item: 

\begin{equation}
h_\xi(\text{\textit{gland}}_2) = g_\xi(f_\xi(\text{\textit{gland}}_2)) = \{ \text{N}, +\text{SG}, -\text{PL}, -\text{M}, +\text{F}, +\text{DEF}, -\text{COL} \}
\end{equation}

\hfill

The preceding operation, which is responsible for the dynamic generation of the template, is warranted by the structural properties of the gradient conditions. Once these gradient conditions have been established, they can be used as parameters of the partial gradient functions $g$, thereby defining a well-formed mapping between items and their corresponding templates, as illustrated below (see equation \eqref{eq5}).

\begin{equation}
\label{eq5}
\begin{split}
\begin{aligned}
h_\xi(\text{\textit{gland}}_2) 	&= g_\xi(f_\xi(\text{\textit{gland}}_2)) \\
						&= \{ \text{N}, +\text{SG}, -\text{PL}, +\text{M}, -\text{F}, +\text{DEF}, -\text{COL} \} \Delta \{ -\text{F}, +\text{M}, +\text{F}, -\text{M} \} \\
						&= \{ \text{N}, +\text{SG}, -\text{PL}, -\text{M}, +\text{F}, +\text{DEF}, -\text{COL} \}
\end{aligned}
\end{split}
\end{equation}

Equation \eqref{eq5} relies on a fundamental property formalised in Lemma \ref{lemma2}, which provides the theoretical basis for the algebraic formulation adopted above.

\begin{lemma} Owing to the fact that the gradient conditions are based on the same operator --- namely the symmetric difference \( \Delta \) --- and given that, in this case, they operate within the framework of the algebra of sets, we can isolate the unknown term $t_j$. Accordingly, they can be generically expressed as: $t_i \,\Delta\,$ $\{-g_1, +g_1, \dots, +g_n,$ $-g_n\} = t_j$. \label{lemma2} \end{lemma} 

\begin{proof} By leveraging the properties of the symmetric difference \citep[see][Chapter~2, Proposition~2.3, p.~10]{GilbertNicholson2004} and assuming \( p, t_i \setminus \{N\}, \) \( t_j \setminus \{N\} \in \mathcal{P}(\Gamma) \) (see Definition~2 in Supplementary Appendix~A), we proceed as follows:
\begin{align*}
t_i \,\Delta\, t_j &= p \\
t_i \,\Delta\, (t_i \,\Delta\, t_j) &= t_i \,\Delta\, p && \text{(Apply } \Delta t_i \text{ to both sides)} \\
(t_i \,\Delta\, t_i) \,\Delta\, t_j &= t_i \,\Delta\, p && \text{(Associativity)} \\
\emptyset \,\Delta\, t_j &= t_i \,\Delta\, p && \text{(Inverse: } t_i \,\Delta\, t_i = \emptyset) \\
t_j &= t_i \,\Delta\, p && \text{(Identity: } \emptyset \,\Delta\, t_j = t_j)
\end{align*}
\end{proof} 

By reformulating the universal gradient condition functions $\mathit{grad cond}_{W}^{n} (f,h; \op, w)$, especially $\op(f(w),t_j)$ (see Definition~\ref{def2}), into a specific algebraic model, we obtain a more direct and transparent method for computing the template of the resultant item. A key advantage of the TBMC model is that the same computational mechanism is used both to learn and to predict. The model learns the parameters of $g$ from the gradient conditions and subsequently applies these learned parameters during prediction. Unlike realizational approaches, where the mappings are typically predefined by rules and learning is absent, the TBMC model integrates parameter acquisition and prediction within a single framework. To conclude this section, we illustrate the derived structure through examples drawn from both stationary and non-stationary states.

\begin{itemize}[labelsep=0mm,parsep=0.3mm,itemsep=0mm,rightmargin=0mm,wide=0pt]
\label{example1}
  \justifying

  \item[\textbf{Example 1}]~Let us compute the resultant template for different synthetic languages with respect to grammatical gender and system state. Conversion will be the only word-formation used in the examples. The equation, $t_i \,\Delta\, p = t_j$, should be interpreted as follows: the left-hand side includes both the template of a base item ($t_i$) and a gradient operand set ($p$), while the right-hand side represents the template of a resultant item ($t_j$).

\begin{itemize}[labelsep=2mm,parsep=0.3mm,itemsep=2mm]

\item[\tiny$\blacktriangleright$] Case of the definiteness-marking synthetic languages:

	\begin{itemize}[labelsep=2mm,parsep=0.3mm,itemsep=2mm]

 \item[$\centerdot$] Gender shift with a masculine base item (e.g., in French: \textit{le sol} ‘ground’ $\rightarrow$ \textit{la sole} ‘sole’)

 \hfill

\hskip -14mm  $\{N,+SG,-PL,+M,-F,+DEF,-COL\}~\Delta~\{-F,+M,+F,-M\}~=~\{N,+SG,-PL,-M,+F,+DEF,-COL\}$

 \hfill

\item[$\centerdot$] Gender shift with a feminine base item (e.g., in French: \textit{la mémoire} ‘memory’ $\rightarrow$ \textit{le mémoire} ‘dissertation’)

\hfill

\hskip -14mm  $\{N,+SG,-PL,-M,+F,+DEF,-COL\}~\Delta~\{-F,+M,+F,-M\}~=~\{N,+SG,-PL,+M,-F,+DEF,-COL\}$

 \hfill

\item[$\centerdot$] Non-gender shift (e.g., in French: \textit{l'hexagone} ‘hexagon’ $\rightarrow$ \textit{l'Hexagone} ‘France’)

\hfill

\hskip -14mm  $ \{N,+SG,-PL,+M,-F,+DEF,-COL\}~\Delta~\emptyset = \{N,+SG,-PL,+M,-F,+DEF,-COL\}$

 \hfill

	\end{itemize}

\end{itemize}

\begin{itemize}[labelsep=2mm,parsep=0.3mm,itemsep=2mm]

\item[\tiny$\blacktriangleright$] Case of the non-definiteness-marking synthetic languages:

	\begin{itemize}[labelsep=2mm,parsep=0.3mm,itemsep=2mm]

 \item[$\centerdot$] Gender shift with a masculine base item (e.g., in Riffian: \textipa{\textit{A-kem:aS}} ‘beam’ $\rightarrow$ \textipa{\textit{DA-kem:aSt}} ‘crutch’)

 \hfill
 
\hskip -14mm   $\{N,+SG,-PL,+M,-F,-COL,+SING\}~\Delta~\{-F,+M,+F,-M\}~=~\{N,+SG,-PL,-M,+F,-COL,+SING\}$

 \hfill

\item[$\centerdot$] Gender shift with a feminine base item (e.g., in Riffian: \textipa{\textit{DA-\textinvscr enZajT}} ‘spoon’ $\rightarrow$ \textipa{\textit{A-\textinvscr enZa}} ‘ladle’)

\hfill

\hskip -14mm  $\{N,+SG,-PL,-M,+F,-COL,+SING\}~\Delta~\{-F,+M,+F,-M\}~=~\{N,+SG,-PL,+M,-F,-COL,+SING\}$

 \hfill

\item[$\centerdot$] Non-gender shift (e.g., in Riffian: \textipa{\textit{a-mhawaD}} ‘act of discussing’ $\rightarrow$ \textipa{\textit{a-mhawaD}} ‘confidante’)

\hfill

\hskip -14mm  $\{N,+SG,-PL,+M,-F,-COL,+SING\}~\Delta~\emptyset~=~\{N,+SG,-PL,+M,-F,-COL,+SING\}$

	\end{itemize}

\end{itemize}

\end{itemize}


\hfill
\\
\hfill

\section{Conclusion}\label{conclusion}

Our theory, grounded in cross-linguistic evidence, proposes a modular approach to noun categories, each associated with an unmarked template. Instead of relying on individual markers, word and meaning formation operates on the entire morphological template. This mechanism is driven by a new computational cognitive component called item-template pairing, and is modelled by the Template-Based and Modular Cognitive model.

This algebraic model can successfully predict the dynamical pairing of items and grammatical markers, including gender markers, which were central to our analysis. Two gradient conditions are defined to explain both gender and non-gender shifts in nouns that occur after word and meaning formation. Our formal demonstration establishes that intra-lexical conversion is formally realisable, constituting a significant theoretical contribution to linguistic theory. 

Through quantitative analysis, we demonstrate that the initial templates of each cognitive set are assigned deterministically; hence, both inter- and intra-lexical derivations follow a predictable behaviour. Consequently, this predictability allows any template shift to be captured by a set parameter, as it has been mathematically proven that the TBMC model possesses the completeness to represent any given shift occurring after word and meaning formation. 

This parsimonious framework offers a unified approach to the varied morphological manifestations characteristic of synthetic languages. By analysing inter-word variations, we have identified the role of modular templates in relation to word and meaning formation. Beyond the formal mathematical expressions, the TBMC model includes multiple core components:

\begin{itemize}
  \item Unmarked modular templates define the grammatical forms that items must pair with.
  \item Modular cognitive sets categorise items and determine their template pairings.
  \item Each modular category has an unmarked template that follows prototypical chains of template shifts, which prevents paradox.
  \item Template shifts occur when transitioning between cognitive sets and are governed by gradient conditions.
\end{itemize}

The aim of this paper was to explore why certain nouns exhibit distinct characteristics across different languages. Our findings demonstrate that these lexemes can be unified within a theoretical framework of morphosyntactic marking. Their shared morphological template suggests that observed differences are strictly paradigmatic rather than structural.

Markedness distinctions among these nouns go beyond gender, encompassing countability as a key factor in shaping unmarked forms, in line with model expectations. Consequently, this theoretical approach provides a unified lens through which to interpret the morphosyntactic diversity present in synthetic languages.

While our focus was on a specific grammatical paradigm, we do not claim to have fully covered the subject. It is important to note that this is just one aspect of a broader investigation. Incorporating additional linguistic units would significantly expand the length of this study. Our contribution serves as a foundation for understanding the underlying patterns of morphosyntax.

Thus, future work could explore the asymmetrical encoding of the feminine marker in synthetic languages. We also believe that the TBMC model could be applied to other linguistic subparts with similar item-template pairings. This mathematical model offers valuable insights and opens venues for future research into non-linear dynamical phenomena in other domains.

\end{Spacing}

\section*{Declarations}

\subsection*{\textbf{\hskip 0.3cm Author contributions}}

The author confirms sole responsibility for the following: study conception and design, data collection, analysis and interpretation of results, and manuscript preparation.

\subsection*{\textbf{\hskip 0.3cm Competing Interests}}
The author declares that he has no conflict of interest.

\subsection*{\textbf{\hskip 0.3cm Ethical approval and consent to participate}}

Not applicable

\subsection*{\textbf{\hskip 0.3cm Funding}}

The author declares that this research did not receive funding.

\subsection*{\textbf{\hskip 0.3cm Availability of data and materials}}

The author confirms that the data supporting the findings of this study are available within the article and attached to the supplementary materials.

\subsection*{\textbf{\hskip 0.3cm Code availability}}

Not applicable

\subsection*{\textbf{\hskip 0.3cm Consent for publication}}

Not applicable

\subsection*{\textbf{\hskip 0.3cm Acknowledgements}}

Not applicable


\end{otherlanguage}

\end{document}


\begin{titlepage}\thispagestyle{empty}\centering
\textbf{SUPPLEMENTARY MATERIALS}\end{titlepage}

\chapter*{Appendix A: Theoretical proposal}

Our model diverges from existing formal language model in several key areas, particularly regarding template encoding and mathematical formalism. This model is based on set-valued set function, which is itself a composition of functions. This model is not designed to manage the pairing of morphemes, but rather to pair items and morphological templates. From a cognitive perspective, the TBMC model is exclusively lexicon-based.

\section{Mathematical foundations of the cognitive model}\label{model_math} 


Our model posits autonomous templates within the lexicon, as suggested by the Word and Paradigm theory. However, the relationship between these templates and lexical items can vary. We employ markedness theory to distinguish between marked and unmarked templates. While other researchers limit markedness to grammatical features, we extend it to include templates.

While all items are mapped to unmarked templates, not all templates have corresponding items. Unpaired templates, even if they contain a single grammatical feature, are considered marked. This perspective views markedness in relation to the image set of a function, where unmarked templates are elements within this set. Items are the pre-images of these unmarked templates. When forming new words, template shifts can occur, leading to a cyclical process where a template can be both marked and unmarked at different times.

Furthermore, any linguistic system may have multiple templates, each with distinct grammatical features. However, items belonging to a specific syntactic category are not randomly assigned to these templates. For a mapping to be valid, it must satisfy two conditions: the number of grammatical features must be identical within each subset, and these subsets must have equivalent grammatical features. This implies a close relationship between templates and syntactic categories.

Definition \ref{def1} introduces the mathematical formulation of these proposals. We will focus on the formal description of templates and item-template pairings.

\begin{deflist}
 \refstepcounter{deflist} \label{def1} 
   \justifying

    \item[Definition 1]\hfill

    Let us consider this definition of the TBMC model of lexicon:

    \begin{itemize}[labelsep=5mm,parsep=0.3mm,itemsep=2mm]
   
      \item[$\bullet$] Let $L$ be the set of the languages
      \item[$\bullet$] Let $W$ be the set of the morphemes or items with $∅ ⊂ W$
      \item[$\bullet$] Let $M$ be the set of the meanings
      \item[$\bullet$] Let $G$ be the set of grammatical meanings with $G ⊂ M$
      \item[$\bullet$] Let $^¬G$ indicate the set $\{$¬g | g $∈ G\}$, that is, $∀$ ¬g, a negative element of $^¬G$ is defined as an element assigned to the empty set of $W$.
      \item[$\bullet$] Let $Γ$ be a set composed of binary grammatical features such that $Γ = G ~\cup$ $^¬G$ with $Γ ⊂ M$. We can write $Γ = \{$+g$_1$, -g$_1$, +g$_2$, -g$_2$, $\ldots$, +g$_n$, -g$_n\}$ with +g$_j$ $∈ G$, -g$_j ∈$ $^¬G$, and $∀$ $j ∈ \mathbb{N}$.
    \end{itemize}

  \end{deflist}

\vskip -5mm

    \begin{deflist}
    \label{def1_1} 
    
    \justifying
    
      \item[Definition 2]  \hfill

      Let us consider this definition of the template:

      \begin{itemize}[labelsep=5mm,parsep=0.3mm,itemsep=2mm]

        \item[$\bullet$] Let $S$ be a set of syntactic categories with  $S ⊂ M$
        \item[$\bullet$] Let $P$ be a set of grammatical paradigms generated by the power-set function $\mathcal{P}(Γ)$ and $P ⊂ \mathcal{P}(Γ)$. Let $Φ$ be sub-collection of $P$.
        \item[$\bullet$] Let $e: Φ → S$ and $∃ϕ_{s∈S} ∈ \{ϕ ~|~ ϕ ∈ Φ, s∈S, e(ϕ)=s\}$ $=$ $e^{-1}(s)$ $⊂ Φ$. The function $e$ holds if only if it satisfies the conditions as follows:

          \begin{alphalist}[labelsep=5mm,leftmargin=10mm,parsep=0.3mm,itemsep=2mm]

            \item If $∀ϕ_{s∈S} ∈ Φ$, its cardinality is equal to a defined number: $\#ϕ=j_{s∈S}$, such that $J = \{j_s\}_{s∈S} \subset \mathbb{N}$  is a set consisting of integer numbers and depending on the set $S$, a function maps $S$ into $J$.

            \item If $∃$ subsets $ϕ_1$, $ϕ_2 ∈ e^{-1}(s)$ $⊂Φ$, such that $ϕ_1 ≠ ϕ_2$ and $e(ϕ_1) = e(ϕ_2)$, then their properties include that the absolute value of $ϕ_1$, $ϕ_2$ are improper subsets of each other: $|ϕ_1| = |ϕ_2|$, and their symmetric difference is an empty set: $|ϕ_1| ~∆~ |ϕ_2|=∅$
            
           

          \end{alphalist}


        \item[$\bullet$] Hence, let $T$ be a family of sets corresponding to the templates such that: $T=\{ϕ \cup \{e_l(ϕ)\} ~|~ ∃ l∈L,∀ ϕ∈Φ_l\}$

      \end{itemize}
      
  \end{deflist}

\vskip -5mm

  \begin{deflist}
      \refstepcounter{subdeflist} 
      \label{def1_2}
      
      \justifying

      \item[Definition 3]  \hfill

      Let us consider this definition of the item-template pairing:

      \begin{itemize}[labelsep=5mm,parsep=0.3mm,itemsep=2mm]

        \item[$\bullet$] Let $V$ be the set of processes of word and meaning’s formation
        \item[$\bullet$] Let $K$ be an unordered set of tuples defined as the set of the retrospective determinants of shifting template such that $K⊆V×T×M×W=$ $\{(v,t,m,w) ~|~$ $v∈V,t∈T,m∈M,w∈W\}$ and the empty subset $∅$ is also an element of $K$
        \item[$\bullet$] Let $f: W → K$ be the backward recursive function that maps an item to an element of $K$, that is each element includes the antecedent shifting factors of an item, and let $g: K → T$ be the gradient function that maps an element of K to the output template. Hence, let $h: W → T$ be the morphosyntactic transfer function defined by the composition function $g∘f$ that maps an item to an output template, such that $∃ l∈L, ∃ w∈W$:

          \hfill

          {\centering
            $h(w)=g(f(w))$\footnote{Languages evolve over time, leading to changes in the mapping between these sets. To represent this, the functions can be expressed as $h^n (w)=g^n (f^n (w))$, where $∀ n∈\mathbb{N}$ indicates different temporal strata.}\par
          }

      \end{itemize}

\end{deflist}


Our model is applicable to all sorts of items that are paired to grammatical templates (such as verbs, nouns, pronouns, etc) and synthetic languages. Its minimalist design, requiring few functions to map grammatical features to items, offers a significant advantage without compromising generality. Thus, lexical entries are encoded as: $(w,h(w)): w∈ W$.

Additionally, the item-template pairing is influenced by a specific meaning encoded in $K$, such that, for $k∈K, ∃m∈k$. These meanings play a role in assigning items to cognitive sets during category changes. Proposition \ref{pro1} formally defines cognitive sets:


\begin{deflist}
  \refstepcounter{newdeflist}
  \refstepcounter{deflist} \label{pro1}
  
  \justifying

    \item[Proposition 1] \hfill

In this context, let us define a cognitive set as the inverse image of a function $f: W \to K$ (see Definition 3). 

    \begin{itemize}[labelsep=5mm,parsep=0.3mm,itemsep=2mm]

      \item[$\bullet$] Let ${\{K_m\}}_{m∈M}$ be an exhaustive disjoint collection of subsets, such that $∃μ,ν ∈ M, μ \neq ν$, and ${K_μ}, {K_ν}  ⊂ K$,  then, ${K_μ} ∩ {K_ν} =  ∅$, and $\bigcup_{m∈ M}  {K_m} = K$
        \item[$\bullet$] Let $\pi_i : K \to M$, $\pi_i(k) = m_i$, be the projection onto the $i$-th coordinate with $i \in \{1,\dots,n\}$.
      \item[$\bullet$] Let $K_m$ be the non-empty subset defined by $\{ k ~|~ k∈ K, \pi_i(k) = m$\} 
      \item[$\bullet$] Hence, let the cognitive set be: $f^{-1}[{K_m}]=\{w ~|~ w∈ W, ∃f(w) \in {K_m}, {K_m} ⊂ K\}$

    \end{itemize}

\end{deflist}

Supplementing Definition \ref{def1}, for expository purposes, let $\mathcal{H}, \mathcal{F}, \mathcal{G}$ be families of partial functions with disjoint domains, where $\mathcal{F} = \{f_i\}_{i \in I}$ and $\mathcal{G} = \{g_i\}_{i \in I}$ for some common index set $I$, then $\mathcal{H} = \mathcal{F} \circ \mathcal{G} = \{f_i \circ g_i \mid \forall f_i \in \mathcal{F}, \forall g_i \in \mathcal{G}\}$ such that, for all $h_i \in \mathcal{H}$, $h_i \colon W \to T$, $\bigcup_{i=1}^{n} \mathrm{dom}\, h_i \subset \mathrm{dom}\, h$, and $\bigcup_{i=1}^{n} \mathrm{ran}\, h_i \subset \mathrm{ran}\, h$. Each of these functions can be interpreted as a prototypical process chain of template shifts. These chains begin with an input head, assuming initial conditions, and demonstrate successive directional derivatives along a specific semantic element.

As an example, consider the partial function $γ$, also denoted as $h_γ$, the Riffian word-form \textit{t-samːeɾθ} is an unmarked pair formed by $γ($\textit{samːeɾ}$_1)=$ $\{N,+SG,-PL,$ $-M,+F,+COL,$ $-SING\}$. We would like to provide another example highlighting a template shift issuing in \textit{samːeɾ}, which derives from \textit{t-samːeɾθ}. The mathematical representation of this word-form is as follows: $f_γ ($\textit{samːeɾ}$_2)=$ $\{CONV,$ $\{N,$ $+SG,$$-PL,$$-M,$ $+F,$$+COL,$$-SING\}$, $C$,  \textit{samːeɾ}$_1\}$, then, from this subset, we obtain the template of \textit{samːeɾ}: $γ($\textit{samːeɾ}$_2)=$ $g_γ (f_γ ($\textit{samːeɾ}$_2))$ $= \{N,+SG,$ $-PL,+M$, $-F,$ $+COL,$ $-SING\}$.